\pdfoutput=1
\documentclass[11pt]{article}

\usepackage{acl}

\usepackage{times}
\usepackage{latexsym}

\usepackage[T1]{fontenc}
\usepackage[utf8]{inputenc}

\usepackage{microtype}
\usepackage{booktabs}
\usepackage{makecell}
\usepackage{multirow}
\usepackage{longtable}
\usepackage{xcolor}
\definecolor{firstplace}{HTML}{FFDDDD} 
\definecolor{secondplace}{HTML}{DDEBFF} 
\usepackage{amsmath}

\usepackage{colortbl}
\usepackage{booktabs} % For professional-looking tables (\toprule, \midrule, \bottomrule)
\usepackage{siunitx}  % For aligning numbers on the decimal point
\usepackage{multirow} % For spanning multiple rows

\usepackage{graphicx}
\usepackage{tikz}
\usepackage{float}
\usepackage{caption}
\usepackage{subcaption}

\usepackage{amssymb}   %
\usepackage{bbding}   
\usepackage{pifont}   

\usepackage{xcolor}
\usepackage{tcolorbox}

\usepackage{tabularx}

\usepackage{algpseudocode}
\usepackage{amsmath}
\usepackage{geometry}
\usepackage[ruled,vlined]{algorithm2e}
\usepackage{amsmath, amssymb}

\algnewcommand\And{\textbf{and}}
\algnewcommand\Or{\textbf{or}}
\algnewcommand\Not{\textbf{not}}

\usepackage{multicol}
\usepackage{blindtext}
\usepackage{dirtree}
\usepackage{chngpage}
\usepackage{minitoc}

\usepackage{url}
\usepackage{hyperref}

\usepackage{tcolorbox}
\tcbuselibrary{skins,breakable}

\usepackage{caption}
\usepackage{subcaption}

\newcommand{\tightEq}[2][]{%
  \par\addvspace{-9pt}\noindent
  \begin{minipage}{\linewidth}
    \centering\small
    \begin{equation}
    \ifx&#1&%
      #2
    \else
      #2\label{#1}%
    \fi
    \end{equation}
    \vspace{-0.5\baselineskip}
  \end{minipage}%
  \par\addvspace{-10pt}
  \noindent
}

\title{See or Say Graphs: Agent-Driven Scalable Graph Structure \\ Understanding with Vision-Language Models}

\author{
    Shuo Han$^{1,2}$\textsuperscript{*} \quad
    Yukun Cao$^{1,2}$\textsuperscript{*} \quad
    Zezhong Ding$^{1,2}$ \\
    \textbf{Zengyi Gao}$^{1,2}$ \quad
    \textbf{S. Kevin Zhou}$^{1,3}$ \quad
    \textbf{Xike Xie}$^{1,2}$\textsuperscript{\textdagger} \\
    $^{1}$University of Science and Technology of China, China \\
    $^{2}$Data Darkness Lab, MIRACLE Center, USTC, China \\
    $^{3}$MIRACLE Center, Suzhou Institute for Advance Research, USTC, China \\
    \texttt{\{shuo.han, ykcho, zezhongding, gzy02\}@mail.ustc.edu.cn} \\
    \texttt{\{skevinzhou, xkxie\}@ustc.edu.cn}
}

\begin{document}
\maketitle
\begin{abstract}
Vision-language models (VLMs) have shown promise in graph structure understanding, but remain limited by input-token constraints, facing scalability bottlenecks and lacking effective mechanisms to coordinate textual and visual modalities. To address these challenges, we propose GraphVista, a unified framework that enhances both scalability and modality coordination in graph structure understanding. For scalability, GraphVista organizes graph information hierarchically into a lightweight GraphRAG base, which retrieves only task-relevant textual descriptions and high-resolution visual subgraphs, compressing redundant context while preserving key reasoning elements. For modality coordination, GraphVista introduces a planning agent that decomposes and routes tasks to the most suitable modality—using the text modality for direct access to explicit graph properties and the visual modality for local graph structure reasoning grounded in explicit topology. Extensive experiments demonstrate that GraphVista scales to large graphs, up to \textbf{200$\times$} larger than those used in existing benchmarks, and consistently outperforms existing textual, visual, and fusion-based methods, achieving up to \textbf{4.4}$\times$ quality improvement over the state-of-the-art baselines by fully exploiting the complementary strengths of both modalities.

\end{abstract}

\section{Introduction} \label{Introduction}

Recently, vision-language models (VLMs) have shown potential for general-purpose graph structure understanding \cite{tang2024graphgpt,kong2025gofa,ding2025msg}, offering new paradigms for solving real-world problems naturally represented as graphs.
A central research focus of this emerging field is to enhance the foundational capability of VLMs to understand graph structures \cite{ren2024survey}, particularly for large-scale graphs, without relying on external tools such as code execution or software systems \cite{ding2025toolcoder,feng2025warriorcoder}. To this end, existing studies have explored two primary graph input modalities for VLMs: textual graph descriptions\footnote{Most early studies \cite{cao-etal-2025-graphinsight,guan2025attention,peng2025rewarding} on text-modality-based graph understanding were developed on LLMs. As VLMs extend LLMs with visual capabilities, this paper presents a unified discussion from the perspective of graph structure understanding with VLMs.}, which convert structural information into natural language for indirect memory and reasoning over node properties and relations \cite{jiang2023structgpt,jin2024graph,zhang2024llm4dyg}; and visual graph representations, which render nodes and edges as structured images to support direct structural perception \cite{li2024visiongraph,zhu-etal-2025-benchmarking}. More recent efforts further combine these modalities to improve structural understanding and complex reasoning on graphs \cite{wei2024gita}.

\setlength{\textfloatsep}{5pt}
 \begin{figure}[t]
	\centering
	\includegraphics[width=1\columnwidth]{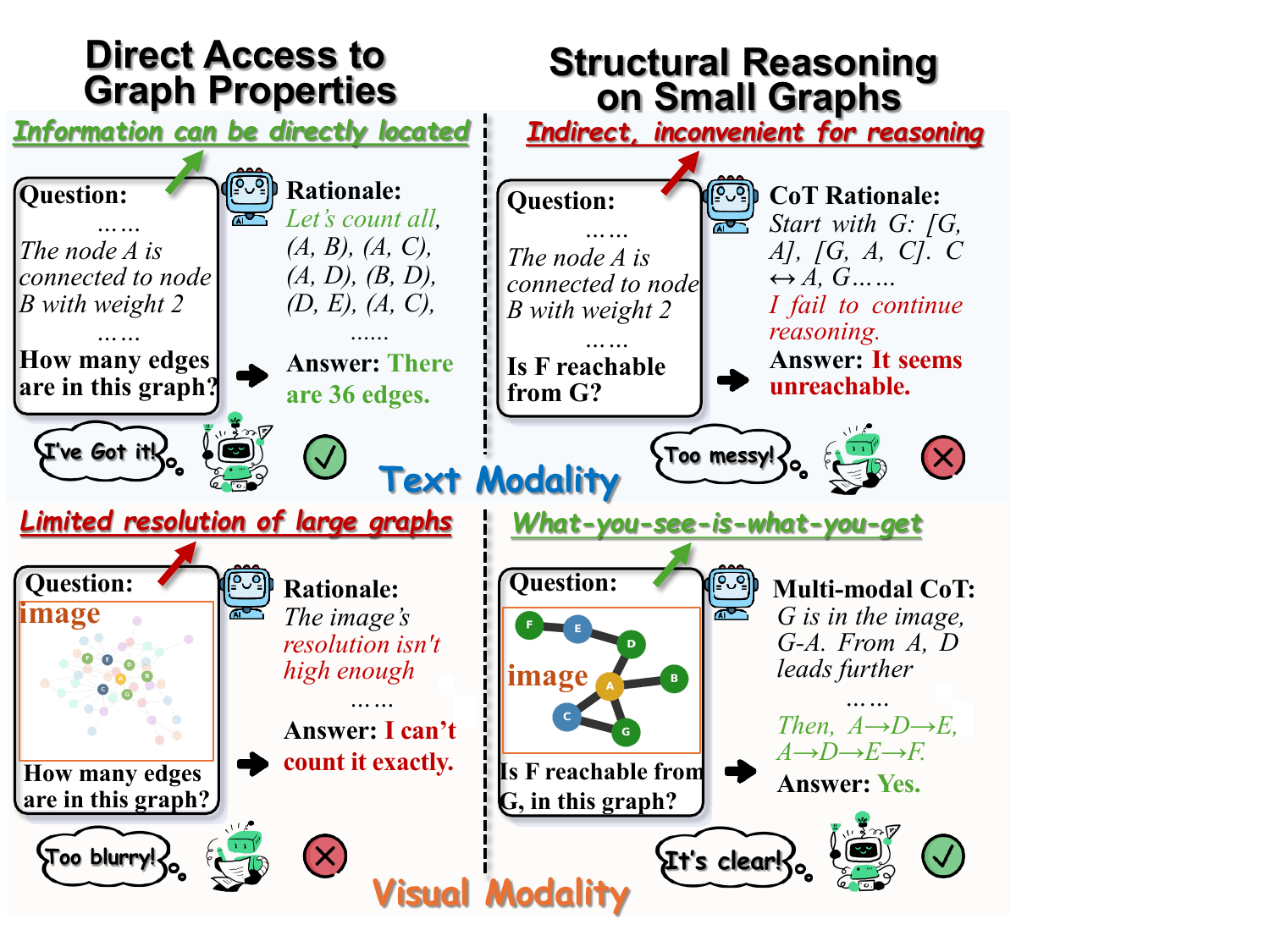}
	% \vspace{-7.5mm}
\caption{Comparison of textual and visual modalities across graph structure understanding tasks}

	\label{fig:intro}
   \vspace{-2pt}
\end{figure}

However, despite these, current VLMs still lack a unified framework that systematically addresses the scalability challenges and the difficulties in coordinating textual and visual modalities inherent in graph structure understanding. Specifically:

\newcommand{\squishlist}{
	\begin{list}{$\bullet$}
		{   \setlength{\itemsep}{0pt}
			\setlength{\parsep}{3pt}
			\setlength{\topsep}{3pt}
			\setlength{\partopsep}{0pt}
			\setlength{\leftmargin}{1.5em}
			\setlength{\labelwidth}{1em}
			\setlength{\labelsep}{0.5em} } }
	\newcounter{Lcount}
	\newcommand{\squishlisttwo}{
		\begin{list}{\arabic{Lcount}. }
			{ \usecounter{Lcount}
				\setlength{\itemsep}{0pt}
				\setlength{\parsep}{0pt}
				\setlength{\topsep}{0pt}
				\setlength{\partopsep}{0pt}
				\setlength{\leftmargin}{2em}
				\setlength{\labelwidth}{1.5em}
				\setlength{\labelsep}{0.5em} } }
		\newcommand{\squishend}{\end{list} }

{\bf Scalability remains a fundamental limitation for both textual and visual graph inputs in VLMs.}
Given their limited input token capacities, textual graph descriptions are constrained by sequence length, making it difficult to encode the full structure of large graphs \cite{chen2024graphwiz,yuan2024gracore}. Visual graph inputs, on the other hand, are restricted by image resolution, resulting in the loss of structural details as graph size increases \cite{zhu-etal-2025-benchmarking}.

{\bf In addition, VLMs face the lack of an effective mechanism to coordinate different input modalities for graph structure understanding,} which prevents them from fully leveraging the distinct strengths and complementary advantages of textual and visual modalities.
As illustrated in Figure~\ref{fig:intro}, under the token length limits of VLMs, the text modality is more effective for tasks that require direct statistical access to explicit graph properties, such as counting nodes, measuring degrees, and assessing connectivity. In these cases, the information can be directly located and reasoned over with minimal effort. By contrast, the visual modality often performs worse because the limited resolution of global visual graphs blurs structural details, causing VLMs to misidentify relevant elements and leading to performance degradation.

For tasks that require local graph structure reasoning\footnote{Tasks based on global graph structure require full-graph access coupled with fine-grained structural reasoning, and can be viewed as a composition of global information retrieval and local structural-reasoning tasks.}, such as shortest-path computation or cycle detection, high-resolution local visual subgraphs offer a clear advantage for VLMs. They provide a ``what-you-see-is-what-you-get'' view of the graph topology and naturally support structural reasoning. In contrast, textual descriptions are often lengthy and lack structural clarity, leaving VLMs with only indirect language cues and thus little structural grounding for effective complex graph reasoning.
%\end{list}

Therefore, a natural conclusion is that fully leveraging VLMs for graph structure understanding requires addressing these two challenges. This entails, on the one hand, preserving as much task-relevant graph information as possible within limited input tokens, and on the other hand, assigning tasks according to modality strengths: letting the text modality efficiently address graph property tasks, while leveraging the visual modality to directly support local structure reasoning tasks.

To this end, we propose {\bf GraphVista}, an agent-driven unified framework designed to coordinate textual and visual modalities for robust graph understanding, ensuring the systematic robustness required to reliably tackle challenging graph tasks.
For the scalability challenges, GraphVista is inspired by the retrieval-augmented generation (RAG) paradigm \cite{lewis2020retrieval}. It organizes graph information hierarchically according to structural importance and stores multi-granularity textual descriptions in a lightweight GraphRAG base. For a given graph structure understanding task, GraphVista retrieves only the relevant information before input, thereby compressing redundant context and preserving key reasoning elements. When necessary, it can also generate high-resolution visual subgraphs from the retrieved local information to support complex reasoning tasks.

To enable effective coordination across modalities, GraphVista is built upon a planning agent that drives the overall workflow. It parses graph tasks, decomposes composite tasks if necessary, and routes them to the appropriate modality. The textual modality branch, enhanced with the RAG techniques, efficiently handles graph properties tasks by retrieving and reasoning over concise textual descriptions. In contrast, the visual modality branch employs the \textbf{Visual Graph Thoughts Agent}, which constructs a multimodal chain-of-thought called Visual Graph Thoughts over high-resolution visual subgraphs, enabling VLMs to perform stepwise visual reasoning grounded in explicit structural evidence for visual tasks.

The main contributions are as follows:
\begin{list}{$\bullet$}
	{   \setlength{\itemsep}{0pt}
		\setlength{\parsep}{1pt}
		\setlength{\topsep}{1pt}
		\setlength{\partopsep}{0pt}
		\setlength{\leftmargin}{1em}
		\setlength{\labelwidth}{1em}
		\setlength{\labelsep}{0.5em} }
    \item We present the first systematic analysis of the scalability challenges and coordination gaps that hinder VLMs in graph structure understanding, providing both conceptual insights and empirical evidence.  

    \item We propose \textbf{GraphVista}, the first unified framework for graph structure understanding with  VLMs, designed around two objectives: improving scalability and enabling coordination across modalities. To this end, GraphVista integrates a set of innovative techniques, such as a \textit{hierarchical GraphRAG base}, a \textit{planning agent} for task routing, and a \textit{multimodal graph thoughts agent} for complex reasoning.  

	\item We present \textbf{Grena}, the first large-scale graph benchmark designed to support step-level evaluation of multimodal graph understanding.
    \item Extensive experiments across diverse graph tasks demonstrate the effectiveness of GraphVista, highlighting its ability to scale to large graphs and to exploit the complementary strengths of textual and visual modalities. 
\end{list}

\vspace{-6pt}
\section{Related Work} \label{Related}
\vspace{-5pt}
\subsection{Text Modality-based Graph Structure Understanding}
\vspace{-5pt}

Most existing studies convert graphs into textual descriptions to enable VLM-based graph structure understanding, primarily focusing on benchmark construction and capability analysis. Recent benchmarks evaluate VLMs on basic graph element comprehension \cite{wang2024can, guo2023gpt4graph,wu2024grapheval2000} and graph-theoretic reasoning tasks \cite{chen2024graphwiz,li2024glbench,tang2024grapharena,luo2024graphinstruct,yuan2024gracore}, indicating that the research in this direction remains in an early exploratory phase. Methodologically, \citet{ge2025sequential,wang2025lost} studied the influence of graph description order, while \citet{cao-etal-2025-graphinsight} identified the “Lost in the Middle” issue in graph sequence. Building upon these findings, \citet{guan2025attention} analyzed VLM attention patterns over graph data, and \citet{peng2025rewarding,zhang2025graphpile} explored task transferability to broader reasoning domains. However, these strategies still face scalability bottlenecks and struggle with complex reasoning tasks.

% For example,
Other studies address graph problems by leveraging external tools. \citet{li2024graphteam,yuan2025ma} employ predefined coding templates for algorithmic reasoning, while \citet{perozzi2024let} trains task-specific GNNs for graph processing. These methods depend on external modules rather than enhancing the intrinsic graph understanding of VLMs, and are thus orthogonal to our work.
\vspace{-5pt}

\subsection{Visual Modality-based Graph Structure Understanding}
\vspace{-5pt}

Visual modality-based graph structure understanding is still in its infancy. \citet{li2024visiongraph}, \citet{zhu-etal-2025-benchmarking}, and \citet{zhao2025underappreciated} benchmark VLMs on graph perception and reasoning, revealing difficulties in capturing structural information. \citet{wei2024gita} further introduces a hybrid multimodal representation directly compatible with VLMs. However, these studies remain empirical and fail to address the scalability, structural understanding, and reasoning limitations inherent to vision-only representations.

\vspace{-6pt}
\section{Methodology} \label{Methodology}
\vspace{-3pt}
\setlength{\textfloatsep}{5pt} %
\begin{figure*}[t]
    \centering
    \includegraphics[width=\textwidth]{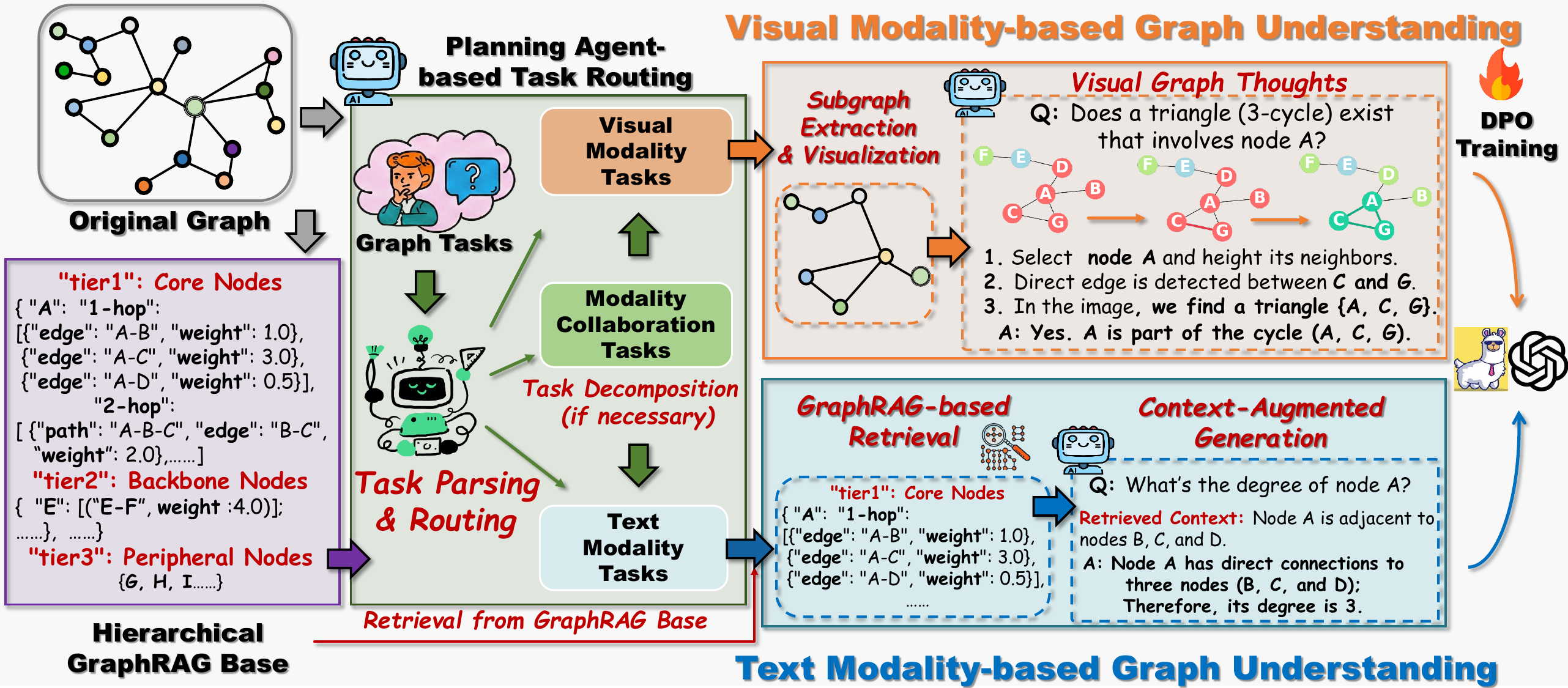}
    \caption{Framework of GraphVista}
    \label{fig:GraphVista}
    \vspace{-15pt}
\end{figure*}

\vspace{-2pt}

\subsection{Overview}\label{Overview}
\vspace{-2pt}

We introduce \textbf{GraphVista}, an agent-driven framework designed to address the challenges of scalability and modality coordination in graph structure understanding\footnote{Aligned with prior studies \cite{wang2024can,chen2024graphwiz,yuan2024gracore}, our work focuses on pure graph structure understanding. The processing of semantic information is orthogonal to our method and can be seamlessly incorporated into GraphVista due to its modular design. }, as shown in Figure~\ref{fig:GraphVista}.

GraphVista is coordinated by a planning agent that directs the entire workflow. Initially, the planning agent constructs a lightweight Hierarchical GraphRAG Base $\mathcal{K}$ by partitioning the graph based on topological centrality. Upon receiving a question, the agent parses the request to identify the task type and key entities, decomposes composite tasks if necessary, and routes them to the appropriate reasoning modality. For text-modality tasks, the text modality-based branch is activated, employing GraphRAG-based retrieval to efficiently extract relevant context from $\mathcal{K}$. For visual-modality tasks, the task is delegated to the visual modality-based branch. In this branch, the Visual Graph Thoughts Agent extracts task-relevant visual subgraphs from $\mathcal{K}$ and performs iterative multimodal reasoning grounded in explicit visual evidence.

The remainder of this section outlines our proposed framework. We begin with the construction of the hierarchical base $\mathcal{K}$ in \S\ref{sec:hierarchical_rag}, followed by the planning agent-driven task routing mechanism in \S\ref{sec:planning_agent}. Finally, we elaborate on the graph understanding modules specific to the visual and textual modalities in \S\ref{sec:vision_understanding} and \S\ref{sec:text_understanding}, respectively.

\vspace{-4pt}
\subsection{Hierarchical GraphRAG Base}
\label{sec:hierarchical_rag}
\vspace{-3pt}

Limited context windows often prevent VLMs from processing complete graphs, leading to the loss of key structural details.  To address this, we propose constructing a local knowledge base $\mathcal{K}$. Diverging from conventional RAG pipelines~\cite{guo2025dior,jin2025hierarchicaldocrefinement} that rely on external knowledge sources, our approach constructs a compact multi-level structural representation directly from the graph and stores it in a lightweight GraphRAG Base, enabling scalable reasoning without external dependencies\footnote{For small graphs ($|V|\leq 15$), we skip the storage phase and feed the entire graph directly into the VLMs.}.

Specifically, we partition the graph $G=(V,E)$ with $|V|$ nodes into three tiers based on the topological centrality of its nodes. Each tier uses a different storage granularity, significantly reducing storage needs while preserving the structural details essential for downstream tasks.

\vspace{-4pt}

\paragraph{Tier 1: Core Nodes.}
This tier comprises nodes with the strongest structural influence, acting as hubs that govern information flow and global graph properties. We employ PageRank, which captures both connectivity and reachability, to identify the top $K_1\%$ of nodes as core nodes. For each, we preserve the 2-hop neighborhood\footnote{We limit the storage scope to a 2-hop radius to optimize the trade-off between structural coverage and computational cost. Extending to 3-hops (or beyond) leads to ``neighbor explosion''~\cite{barabasi1999emergence, GrabowGKT15}, yet remote nodes contribute little to local structural reasoning~\cite{LiHW18, Liben-NowellK03}. We adopt this generalized strategy to ensure broad applicability; advanced GraphRAG storage optimizations are orthogonal to our contribution and can be seamlessly integrated.} $\mathcal{N}_2(v)$ to explicitly retain fine-grained local topology.

\vspace{-3pt}

\paragraph{Tier 2: Backbone Nodes.} 
Situated along inter-community paths, these nodes are critical for maintaining global connectivity. We select them using Betweenness Centrality, prioritizing nodes that frequently bridge shortest paths between pairs. The subsequent $K_2\%$ of nodes are retained; however, to balance representational fidelity with storage efficiency, we store only the 1-hop neighborhood $\mathcal{N}_1(v)$ for each backbone node.
\vspace{-3pt}

\paragraph{Tier 3: Peripheral Nodes.} 
    The remaining nodes are classified as peripheral, whose structural role is primarily defined through their links to higher-tier nodes. To minimize redundancy, we adopt a conditional storage policy: the 1-hop neighborhood $\mathcal{N}_1(v)$ of a peripheral node is stored only if it is not connected to any Tier 1 or Tier 2 node. Otherwise, its structural information is implicitly covered by the neighborhoods of higher-tier nodes, ensuring maximal storage efficiency\footnote{In practice, $\mathcal{K}$ can be configured to utilize only a subset of tiers (e.g., only Tier 1) to adapt to storage or computational constraints, providing flexibility across application scenarios.}.

$\mathcal{K}$ not only addresses scalability but also establishes the information backbone for subsequent multimodal collaborative reasoning. In the text modality-based branch, it enables efficient retrieval of relevant structured context, while in the visual modality-based branch, it guides the precise extraction of local visual subgraphs.

\vspace{-4pt}

\subsection{Planning Agent-based Task Routing}\label{sec:planning_agent}
\vspace{-2pt}

This module employs a planning agent to allocate each task to the optimal reasoning modality, either textual or visual.

\vspace{-3pt}

\paragraph{Task Parsing.} Given a natural language question $Q$ over $G$, the planning agent employs semantic parsing to determine the task type $T$ and extract relevant entities $\mathcal{E}$:
\tightEq[eq:planningparse]{(T, \mathcal{E}) = f_{\text{parse}}(Q, G)}
, where $f_{\text{parse}}$ utilizes few-shot prompting\footnote{The planning agent first matches the question against predefined task templates; if unmatched, it falls back to semantic parsing to infer the intent and choose the reasoning modality. Implementation details and a quantitative analysis of task classification error rates are provided in Appendix~\ref{sec:task_parsing_routing_details}.} to infer task semantics, producing a task category $T$ (e.g., neighbor identification) and a set of entity pairs $\mathcal{E}$. To ensure consistent downstream processing, queries involving a single entity (e.g., node degree) are normalized to $(v_i, \emptyset)$ using a null placeholder.
\vspace{-3pt}

\paragraph{Task Routing and Execution Planning.}
{After parsing the task, the planning agent routes the problem to the appropriate processing module based on the task type $T$ and formulates a detailed execution plan. 
Integrating task designs from existing graph understanding benchmarks, the planning agent routes tasks based on their dependence on structural reasoning, thereby aligning each task with the respective strengths of text and vision modalities\footnote{Composite tasks that can be decomposed into single-modality operations (e.g., determining which of two nodes shares more common neighbors with a given node) can be handled by the planning agent through sequential execution of the corresponding subtasks.}. Specifically, tasks that do not require structural reasoning are handled by the text modality, tasks requiring local structural reasoning are assigned to the visual modality, and tasks involving global graph structure are decomposed into stepwise text- and vision-based subtasks.}

\vspace{-4pt}

\begin{tcolorbox}[colback=white, colframe=black, width=\linewidth, boxrule=0.4pt, rounded corners, left=1mm, right=1mm, top=1mm, bottom=1mm, boxsep=1mm]
    \textbf{Text-Modality Task Examples:}
    
    \textit{“What is the total number of nodes in this graph?”}; \textit{“What is the degree of node 15?”}
    
    \vspace{1.5mm}
    \textbf{Visual-Modality Task Examples:}
    
    \textit{“Find the shortest path between nodes A and B.”}; 
    \textit{“Whether node A is part of any triangle.”}
    
    \vspace{1.5mm}
    \textbf{Modality-Collaborative Task Examples:}
    
    \textit{“Calculate the diameter of the graph.” }
    
\textit{\textbf{Decomposition}: (i) Identify peripheral nodes using text; (ii) Estimate pairwise distances through visual subgraph reasoning.”}
\end{tcolorbox}

\vspace{-8pt}

\vspace{-3pt}

\paragraph{Text-Modality Tasks.}

These tasks do not require structural reasoning over the graph. Instead, they rely on direct access to individual nodes or edges for basic statistical retrieval, such as counting the total number of nodes or querying the degree of a given node.
Accordingly, the planning agent directs these tasks to the GraphRAG agent module, providing a retrieval strategy for execution over $\mathcal{K}$.

\vspace{-3pt}
\paragraph{Visual-Modality Tasks.}
These tasks involve reasoning over local graph structures rather than simple retrieval. They focus on topological relationships within a limited subgraph (e.g., shortest-path tasks). In such cases, visual subgraph representations provide a clearer and more efficient form for structural reasoning than textual descriptions.
Consequently, the planning agent delegates such tasks to the visual graph thoughts agent, while decomposing the task into algorithm-guided substeps and constructing a high-level reasoning plan.

\vspace{-3pt}
 \paragraph{Modality-Collaborative Tasks.}

These tasks rely on global graph structure understanding that combines full-graph access with fine-grained structural analysis, which is often beyond a single modality. They are therefore formulated as a sequence of steps that alternate between global information access and local structural reasoning. The planning agent decomposes the task into sequential text- and visual-modality sub-tasks.

\vspace{-7pt}

\subsection{Visual Modality-based Graph Structure Understanding} \label{sec:vision_understanding}

\vspace{-4pt}
This module leverages visual-modality graph representations to handle visual-modality tasks that require fine-grained topological understanding and multi-step logical inference. To address scalability, we extract a task-relevant $k$-hop subgraph $G'=(V',E')$ centered on key entities $\mathcal{E}$, with at most $N_{\max}$ nodes (as detailed in \textbf{Appendices~\ref{appendix_Subgraph_Ext} and \ref{sec:appendix_c}}). $G'$ is visualized as $G_{\text{image}} = f_{\mathrm{viz}}(G')$ via a task-driven strategy that adapts layouts to the task and simplifies dense subgraphs for visual clarity, ensuring that essential structural information remains perceivable to the VLM.

\vspace{-6pt}
\subsubsection{Visual Graph Thoughts}
\vspace{-5pt}
Existing text-based CoT methods often suffer from \textit{pseudo-visual reasoning}, in which visual information gradually fades as the reasoning chain lengthens \cite{chen2024m3cot,zhang2024multimodal,zou2024vgbench,cheng2025comt}. To overcome this limitation in multimodal reasoning, we introduce Visual Graph Thoughts.

After obtaining the visual subgraph $G_{\text{image}}$, the visual reasoning agent $\mathcal{M}_{\text{VRA}}$, instantiated from a VLM, performs iterative multimodal reasoning under the guidance of a high-level plan $\Pi = (\pi_1, \pi_2, \dots, \pi_n)$ generated by the planning agent.

We formalize this process as a state-transition system. At step $t$ ($1 \le t \le n$), the reasoning state $S_t$ is defined by the current visual representation $G_{\text{image}}^{(t-1)}$, the accumulated reasoning history $H_{t-1}$, and the current plan instruction $\pi_t$. Conditioned on $S_t$, $\mathcal{M}_{\text{VRA}}$ produces an intermediate reasoning output $o_t$ and an action $a_t$:
\tightEq[eq:reasonstep]{(o_t, a_t) = \mathcal{M}_{\text{VRA}}(S_t) = \mathcal{M}_{\text{VRA}}(G_{\text{image}}^{(t-1)}, H_{t-1}, {\pi}_t).}
If $a_t$ invokes $f_{\mathrm{viz}}$ (e.g., node highlighting), the visual representation is updated as $G_{\text{image}}^{(t)} = f_{\mathrm{viz}}(G_{\text{image}}^{(t-1)}, a_t)$; otherwise, $G_{\text{image}}^{(t)} = G_{\text{image}}^{(t-1)}$. This closed-loop process continues until all plan steps are executed. Finally, $\mathcal{M}_{\text{VRA}}$ aggregates the reasoning trace $H_n$ to produce the final answer.

\vspace{-3pt}
\subsubsection{Aligning Visual Graph Thoughts with Process-level DPO}

\vspace{-3pt}

To further improve the multimodal reasoning capability of $\mathcal{M}_{\text{VRA}}$ by promoting the generation of reliable visual graph thoughts, we adopt the process-level Direct Preference Optimization (DPO) training strategy \cite{rafailov2023direct}, which models the full multimodal reasoning trajectory and provides stable, step-level supervision for graph inference. Advanced reinforcement learning methods \cite{icml/Hu0Q0CW025,dinucu-jianu-etal-2025-teaching,corr/abs-2503-14476} are orthogonal to our framework and can be naturally integrated. We adopt the standard process-level DPO formulation to ensure generality and training stability.

\vspace{-2pt}
\paragraph{1. Construction of Process Preference Dataset.}
\vspace{-1pt}
We construct a process preference dataset $\mathcal{D}$, where each sample is a triplet $(\mathbf{x}, y_w, y_l)$. Here, $\mathbf{x}$ denotes the multimodal input, $y_w$ represents the ``Chosen'' path (preferred reasoning trajectory), and $y_l$ represents the ``Rejected'' path (non-preferred trajectory). The construction details are in \textbf{Appendix \ref{Benchmark}}.
\vspace{-4pt}
\paragraph{``Chosen'' Path ($y_w$).} We construct standardized reasoning templates aligned with the logical structure of each graph task. Based on these templates, we generate ground-truth reasoning steps and answers using large-scale synthetic graphs (e.g., ER and BA graphs \cite{li2005towards}) and validate their correctness through graph analysis libraries.
\vspace{-2pt}
\paragraph{``Rejected'' Path ($y_l$).}
\vspace{-2pt}
To help the VLM distinguish between preferred and non-preferred paths, we construct negative samples $y_l$ by modifying $y_w$. Specifically, we extend error construction strategies from existing benchmarks \cite{chen2024graphwiz,yuan2024gracore} to multimodal reasoning tasks. The error categories are summarized in \textbf{Table~\ref{tab:error_categories}}.

\vspace{-4pt}

\paragraph{2. DPO Training Process.}
To align the generation of visual graph thoughts with ground-truth multimodal reasoning trajectories, we fine-tune $\mathcal{M}_{\text{VRA}}$ using DPO on $\mathcal{D}$. For stability, we use a reference policy $\pi_{\text{ref}}$, defined as a copy of $\mathcal{M}_{\text{VRA}}$ that has been supervised to fine-tune in $\mathcal{D}$ and kept fixed during DPO training. The DPO objective minimizes the following loss function:

\tightEq{\mathcal{L}_{\mathrm{DPO}} =
-\mathbb{E}_{(\mathbf{x}, y_w, y_l) \sim \mathcal{D}}
\left[\log \sigma \!\left(\beta \log \frac{r_{\theta}(y_w|\mathbf{x})}{r_{\theta}(y_l|\mathbf{x})}\right)\right]}
\vspace{4pt}
, where the policy ratio is defined as $r_{\theta}(y|\mathbf{x}) = \pi_{\theta}(y|\mathbf{x}) / \pi_{\mathrm{ref}}(y|\mathbf{x})$, and $\beta$ controls the regularization strength. Maximizing the log-probability margin between preferred ($y_w$) and rejected ($y_l$) trajectories effectively suppresses predefined reasoning errors. The resulting policy generates visual graph thoughts that are logically coherent and faithfully grounded in visual evidence. Training details are available in \textbf{Appendices~\ref{sec:experimental_settings} and ~\ref{sec:stability_DPO}}.

\vspace{-5pt}

\subsection{Text Modality-based Graph Structure Understanding}\label{sec:text_understanding}
\vspace{-4pt}

For text-modality tasks, we adopt GraphRAG-based retrieval for targeted information extraction. Note that existing RAG variants and optimizations~\cite{deng2023regavae, lee2025hybgrag, gao2025frag, chang2025mainrag} are orthogonal to GraphVista and can be easily integrated to improve retrieval quality. To maintain generality, we employ the standard RAG formulation in this work.

Given the parsed tuple $(Q, \mathcal{E})$, the relevant context $\mathcal{C}$ is extracted from $\mathcal{K}$ by aggregating the structural descriptions of entities in $\mathcal{E}$. The VLM then generates the answer conditioned on both $Q$ and $\mathcal{C}$.

\vspace{-5pt}
\section{Evaluation}
\label{Experiment}
\vspace{-2pt}

\setlength{\heavyrulewidth}{0.35pt}
\setlength{\lightrulewidth}{0.29pt}
\setlength{\arrayrulewidth}{0.29pt}

\begin{table*}[t]
\setlength{\abovecaptionskip}{0pt}
\setlength{\belowcaptionskip}{0pt}
\setlength{\tabcolsep}{6pt}
\renewcommand{\arraystretch}{0.48}
\centering
\caption{
Performance comparison on Grena and GraphSQA Benchmark.
{\colorbox{firstplace}{\color{red}\bf Red}} and {\colorbox{secondplace}{Blue}}
highlight the best and second-best results, respectively.
}
\label{tab:overall_res}
\small
\resizebox{\textwidth}{!}{%
\begin{tabular}{l l|cccc @{\hspace{1.5em}} ccc}
\toprule
\multirow{2}{*}{\textbf{Method}} & \multirow{2}{*}{\textbf{Model}} 
& \multicolumn{4}{c}{\textbf{Grena}} 
& \multicolumn{3}{c}{\textbf{GraphSQA}} \\
\cmidrule(lr){3-6} \cmidrule(lr){7-9}
& & \textbf{Text} & \textbf{Visual} & \textbf{Collab.} & \textbf{Overall}
& \textbf{Text} & \textbf{Visual} & \textbf{Overall} \\
\midrule
% -----------------------------
\multirow{6}{*}{Text-only}
 & GLM-4.1V-9B       & 0.0352 & 0.0055 & 0.0110 & 0.0217 & 0.2532 & 0.1670 & 0.1929 \\
 & InternVL3-9B      & 0.0452 & 0.0180 & 0.0303 & 0.0345 & 0.4788 & 0.2498 & 0.3185 \\
 & Qwen2.5-VL-7B     & 0.0558 & 0.0345 & 0.0337 & 0.0450 & 0.2340 & 0.1122 & 0.1487 \\
 & Qwen3-8B          & 0.0403 & 0.0367 & 0.0096 & 0.0321 & 0.1351 & 0.1352 & 0.1352 \\
 & Gemma-3-12B       & 0.1246 & 0.0591 & 0.0337 & 0.0858 & 0.5917 & 0.3245 & 0.4047 \\
 & GPT-5-mini        & 0.1451 & 0.0933 & 0.0515 & 0.1093 & 0.6464 & 0.6082 & 0.6197 \\
\midrule
\multirow{3}{*}{GraphPRM (DPO)}
 & Qwen2.5-VL-7B (DPO) & 0.0589 & 0.0410 & 0.0314 & 0.0477 & 0.7198 & 0.3455 & 0.4578 \\
 & Qwen3-8B (DPO)      & 0.0446 & 0.0379 & 0.0068 & 0.0339 & 0.3629 & 0.2357 & 0.2739 \\
 & Gemma-3-12B (DPO)   & 0.1281 & 0.0590 & 0.0434 & 0.0899 & 0.6235 & 0.3455 & 0.4289 \\
\midrule
\multirow{5}{*}{GraphInsight}
 & GLM-4.1V-9B       & 0.1951 & 0.2655 & 0.2172 & 0.2189 & 0.4313 & 0.3273 & 0.3585 \\
 & InternVL3-9B      & 0.2507 & 0.2915 & 0.2088 & 0.2515 & 0.7564 & 0.2315 & 0.3890 \\
 & Qwen2.5-VL-7B     & 0.2011 & 0.1078 & 0.0976 & 0.1520 & 0.5769 & 0.2293 & 0.3336 \\
 & Gemma-3-12B       & 0.2469 & 0.2357 & 0.1131 & 0.2123 & 0.7372 & 0.2478 & 0.3946 \\
 & GPT-5-mini        & 0.3031      & 0.3429      &  0.2843      & 0.3091      & 0.7211 & 0.6446 & 0.6676 \\
\midrule
\multirow{3}{*}{GraphToken}
 & Qwen2.5-VL-7B     & 0.0141 & 0.0982 & 0.0396 & 0.0423 & 0.0388 & 0.1770 & 0.1355 \\
 & Qwen3-8B          & 0.0052 & 0.0665 & 0.0089 & 0.0222 & 0.1742 & 0.1508 & 0.1578 \\
 & Gemma-3-12B       & 0.0191 & 0.1064 & 0.0475 & 0.0488 & 0.0566 & 0.1091 & 0.0933 \\
\midrule
\multirow{4}{*}{GITA}
 & GLM-4.1V-9B       & 0.0665 & 0.2152 & 0.1835 & 0.1333 & 0.3397 & 0.1562 & 0.2112 \\
 & InternVL3-9B      & 0.0714 & 0.1045 & 0.1546 & 0.0998 & 0.3910 & 0.1755 & 0.2401 \\
 & Qwen2.5-VL-7B     & 0.1099 & 0.1799 & 0.2271 & 0.1561 & 0.3205 & 0.1596 & 0.2079 \\
 & Gemma-3-12B       & 0.1218 & 0.1591 & 0.2344 & 0.1583 & 0.2692 & 0.1739 & 0.2025 \\
\midrule
\multirow{5}{*}{GraphVista (Ours)}
 & GLM-4.1V-9B
   & 0.9352 & 0.3540 & 0.2559 & 0.6214
   & 0.7954 & 0.6657 & 0.7046 \\
 & InternVL3-9B
   & 0.9357 & 0.3701 & 0.3449 & 0.6469
   & 0.7970 & \cellcolor{secondplace}{0.6886} & \cellcolor{secondplace}{0.7211} \\
 & Qwen2.5-VL-7B (DPO)
   & \cellcolor{secondplace}{0.9363}
   & 0.4309
   & 0.4478
   & 0.6876
   & 0.7973
   & 0.6806
   & 0.7156 \\
 & Gemma-3-12B (DPO)
   & 0.9361
   & \cellcolor{secondplace}{0.4431}
   & \cellcolor{secondplace}{0.6018}
   & \cellcolor{secondplace}{0.7272}
   & \cellcolor{secondplace}{0.7991}
   & 0.6815
   & 0.7168 \\
 & GPT-5-mini
   & \cellcolor{firstplace}{\bf\color{red}0.9406}
   & \cellcolor{firstplace}{\bf\color{red}0.4526}
   & \cellcolor{firstplace}{\bf\color{red}0.6241}
   & \cellcolor{firstplace}{\bf\color{red}0.7372}
   & \cellcolor{firstplace}{\bf\color{red}0.8497}
   & \cellcolor{firstplace}{\bf\color{red}0.7967}
   & \cellcolor{firstplace}{\bf\color{red}0.8126} \\
\bottomrule
\end{tabular}%
}
\vspace{-19pt}
\end{table*}

\subsection{Experimental Setup}
\label{sec:exp_setup}
\vspace{-2pt}

\paragraph{Baselines.}
\label{sec:exp_baselines}
We compare GraphVista with state-of-the-art baselines across three categories:
\textbf{(a) Text-based methods}, which reason exclusively over textual descriptions, including the standard Text-only baseline, GraphPRM~\cite{peng2025rewarding} (trained via DPO), and GraphInsight~\cite{cao-etal-2025-graphinsight}.
\textbf{(b) GNN-based methods}, such as GraphToken~\cite{perozzi2024let}, which encode structural information into latent representations\footnote{These methods focus on graph encoding rather than VLM intrinsic understanding. We include them for completeness.}.
\textbf{(c) Hybrid methods}, represented by GITA~\cite{wei2024gita}, which integrate both visual and textual modalities.

\vspace{-3pt}
\paragraph{Benchmarks and Evaluation Metrics.} 
\vspace{-2pt}

\label{sec:exp_benchmarks}

Current graph understanding benchmarks suffer from restricted scale and topological diversity, with insufficient support for multi-step visual reasoning. We introduce \textbf{Grena}, a large-scale benchmark designed to evaluate VLMs across 20 distinct task types, covering the scope of existing baselines~\cite{wang2024can,yuan2024gracore,zhu-etal-2025-benchmarking}. Grena spans a wide range of graph scales (up to 2,050 nodes) and topologies. Further details are provided in \textbf{Appendix~\ref{Benchmark}}.

We use Grena to evaluate graph structure understanding performance on large-scale graphs ($|V| \in [50, 2050]$), covering a total of 22,800 tasks. Complementarily, we use the GraphSQA benchmark \cite{cao-etal-2025-graphinsight} for small-graph structure understanding, with $|V|$ in the range of $[15, 50]$\footnote{As this small-graph benchmark lacks modality-collaborative tasks, we do not separate local and global tasks. Instead, we directly adopt its original classification, referring to the two categories as ``Text'' and ``Visual''.}. Across all tasks and benchmarks, we use \textbf{Accuracy (ACC)} as the evaluation metric. Methods requiring DPO are trained on $\mathcal{D}$, as detailed in \textbf{Appendix~\ref{sec:experimental_settings}}.

\begin{table}[t]
\setlength{\belowcaptionskip}{0pt}
\renewcommand{\arraystretch}{0.86}
\centering
\caption{Ablation study of reasoning strategies in GraphVista. Best results within each VLM group are bolded, and second-best results are underlined.}
\label{tab:ablation_cot}
\resizebox{\columnwidth}{!}{%
\begin{tabular}{lcccc}
\toprule
\textbf{Strategy} & \textbf{Text} & \textbf{Visual} & \textbf{Collab.} & \textbf{Overall} \\
\midrule
% --- GLM-4.1V-9B ---
\rowcolor{gray!20}
\multicolumn{5}{l}{\textit{GLM-4.1V-9B}} \\
CoT (Baseline)                 
& \multirow{3}{*}{0.9352} 
& 0.1376 
& 0.1667 
& 0.5433 \\
Visual Graph Thoughts (Text)   
&                         
& \underline{0.2734} 
& \underline{0.1902} 
& \underline{0.5846} \\
Visual Graph Thoughts (Visual) 
&                         
& \textbf{0.3540} 
& \textbf{0.2559} 
& \textbf{0.6214} \\
\midrule
% --- InternVL3-9B ---
\rowcolor{gray!20}
\multicolumn{5}{l}{\textit{InternVL3-9B}} \\
CoT (Baseline)                 
& \multirow{3}{*}{0.9357} 
& 0.1422 
& 0.1332 
& 0.5368 \\
Visual Graph Thoughts (Text)   
&                         
& \underline{0.2879} 
& \underline{0.2511} 
& \underline{0.6031} \\
Visual Graph Thoughts (Visual) 
&                         
& \textbf{0.3701} 
& \textbf{0.3449} 
& \textbf{0.6469} \\
\midrule
% --- Qwen2.5-VL-7B (DPO) ---
\rowcolor{gray!20}
\multicolumn{5}{l}{\textit{Qwen2.5-VL-7B (DPO)}} \\
CoT (Baseline)                 
& \multirow{3}{*}{0.9363} 
& 0.1691 
& 0.1684 
& 0.5525 \\
Visual Graph Thoughts (Text)   
&                         
& \underline{0.3629} 
& \underline{0.3355} 
& \underline{0.6431} \\
Visual Graph Thoughts (Visual) 
&                         
& \textbf{0.4309} 
& \textbf{0.4478} 
& \textbf{0.6876} \\
\midrule
% --- Gemma-3-12B (DPO) ---
\rowcolor{gray!20}
\multicolumn{5}{l}{\textit{Gemma-3-12B (DPO)}} \\
CoT (Baseline)                 
& \multirow{3}{*}{0.9361} 
& 0.1676 
& 0.1703 
& 0.5525 \\
Visual Graph Thoughts (Text)   
&                         
& \underline{0.3551} 
& \underline{0.4200} 
& \underline{0.6610} \\
Visual Graph Thoughts (Visual) 
&                         
& \textbf{0.4431} 
& \textbf{0.6018} 
& \textbf{0.7272} \\
\bottomrule
\end{tabular}%
}
\vspace{-5pt}

\end{table}

\vspace{-3pt}
\paragraph{Models.}\label{sec:exp_models}
\vspace{-2pt}

We instantiate all methods using recent open-source, competitive VLMs, including Gemma-3-12B~\cite{team2025gemma}, Qwen2.5-VL-7B~\cite{bai2025qwen2}, InternVL3-9B~\cite{zhu2025internvl3}, and GLM-4.1V-9B-Thinking~\cite{hong2025glm}, a reasoning-oriented VLM. For text-based methods, we include more capable LLMs such as Qwen3-8B \cite{yang2025qwen3} to ensure a fairer comparison. For training-free methods, we include GPT-5-mini to assess generality.

\begin{table}[t]

\centering

\caption{Ablation comparing frozen base models (``Frozen'') and DPO-fine-tuned models (``DPO''). {\colorbox{firstplace}{\color{red}\bf Red}} and {\colorbox{secondplace}{Blue}} denote the best and second-best results.}
\label{tab:DPO}
\small
\resizebox{\columnwidth}{!}{
\begin{tabular}{l|l|cccc}
\toprule
\textbf{Model} & \textbf{Strategy} & \textbf{Text} & \textbf{Visual} & \textbf{Collab.} & \textbf{Overall} \\
\midrule

% --- Qwen2.5-VL-7B ---
\multirow{2}{*}{Qwen2.5-VL-7B}
 & Frozen & \multirow{2}{*}{0.9363} & 0.3951 & 0.3589 & 0.6571 \\
 & DPO    &                        & \cellcolor{secondplace}{0.4309} & 0.4478 & 0.6876 \\
\midrule

% --- Gemma-3-12B ---
\multirow{2}{*}{Gemma-3-12B}
 & Frozen & \multirow{2}{*}{0.9361} & 0.4005 & \cellcolor{secondplace}{0.5387} & \cellcolor{secondplace}{0.7010} \\
 & DPO    &                        & \cellcolor{firstplace}{\bf\color{red}0.4431} 
                                  & \cellcolor{firstplace}{\bf\color{red}0.6018} 
                                  & \cellcolor{firstplace}{\bf\color{red}0.7272} \\
\bottomrule
\end{tabular}}
\vspace{-4pt}
\end{table}

\begin{figure}[t]
	\centering

	\begin{subfigure}[b]{0.234\textwidth}
		\centering
		\includegraphics[width=\textwidth]{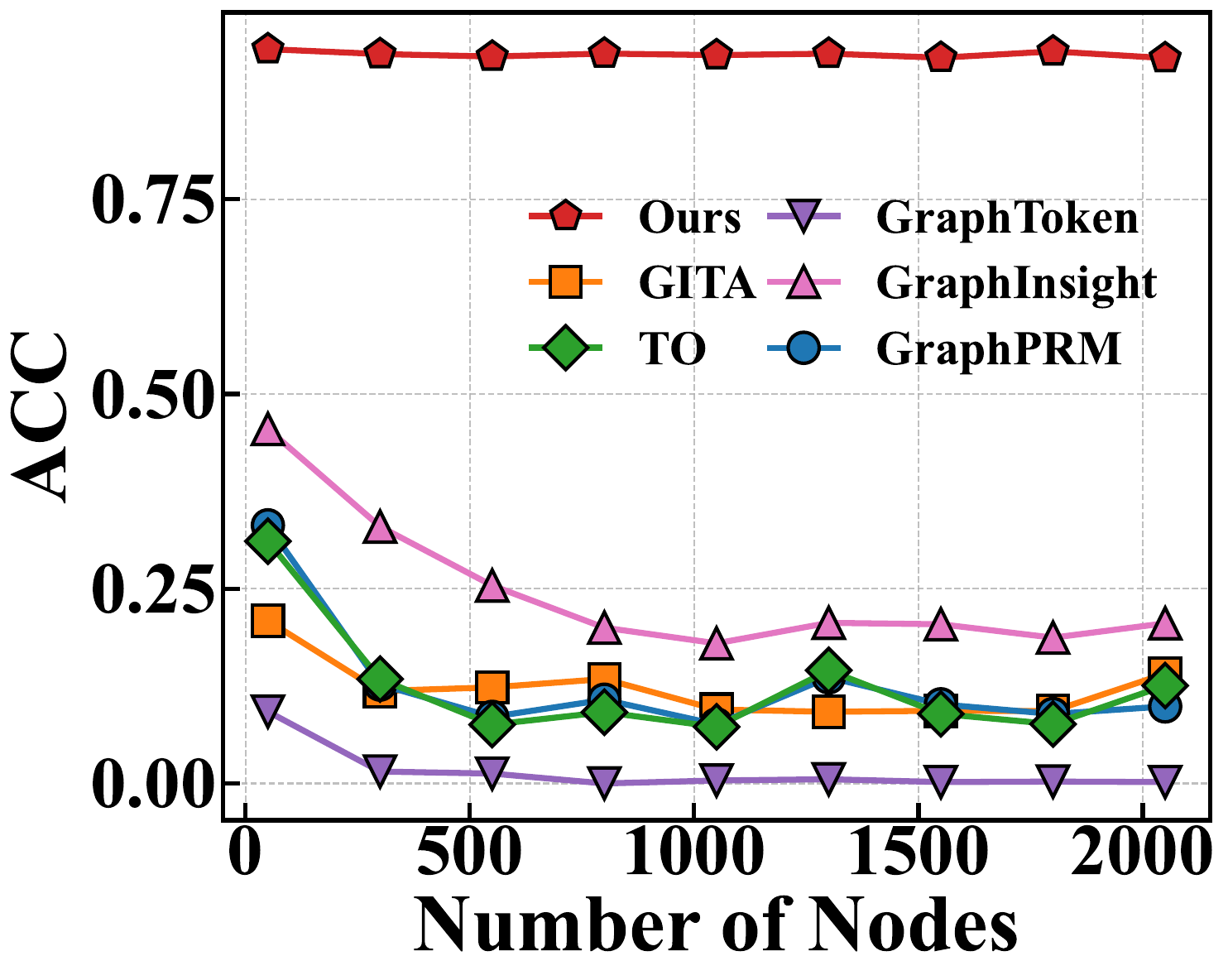}
		\caption{Text (Gemma-3)}
		\label{fig:gemma_text}
	\end{subfigure}
	\hfill
	\begin{subfigure}[b]{0.234\textwidth}
		\centering
		\includegraphics[width=\textwidth]{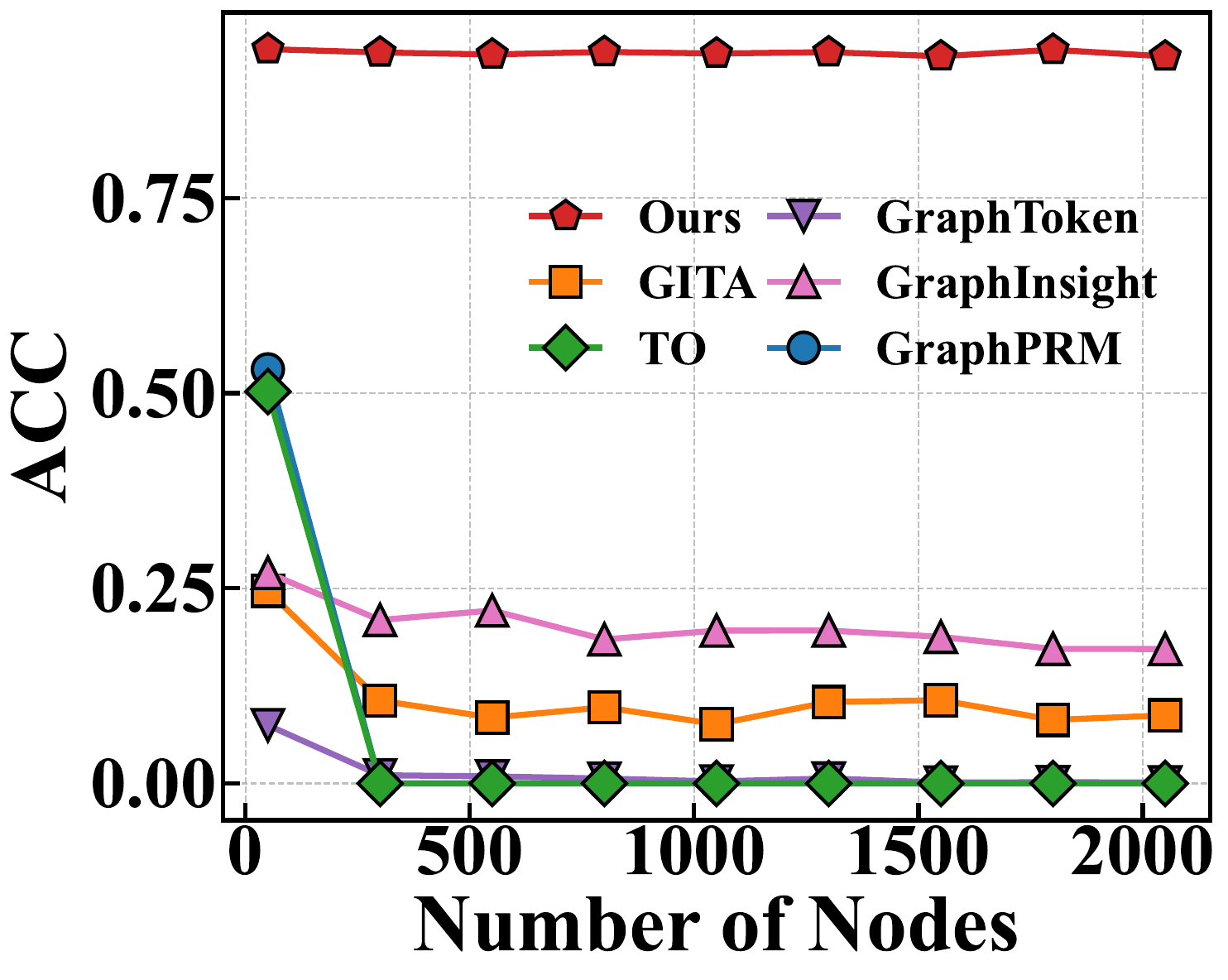}
		\caption{Text (Qwen2.5-VL)}
		\label{fig:qwen_text}
	\end{subfigure}

	\begin{subfigure}[b]{0.234\textwidth}
		\centering
		\includegraphics[width=\textwidth]{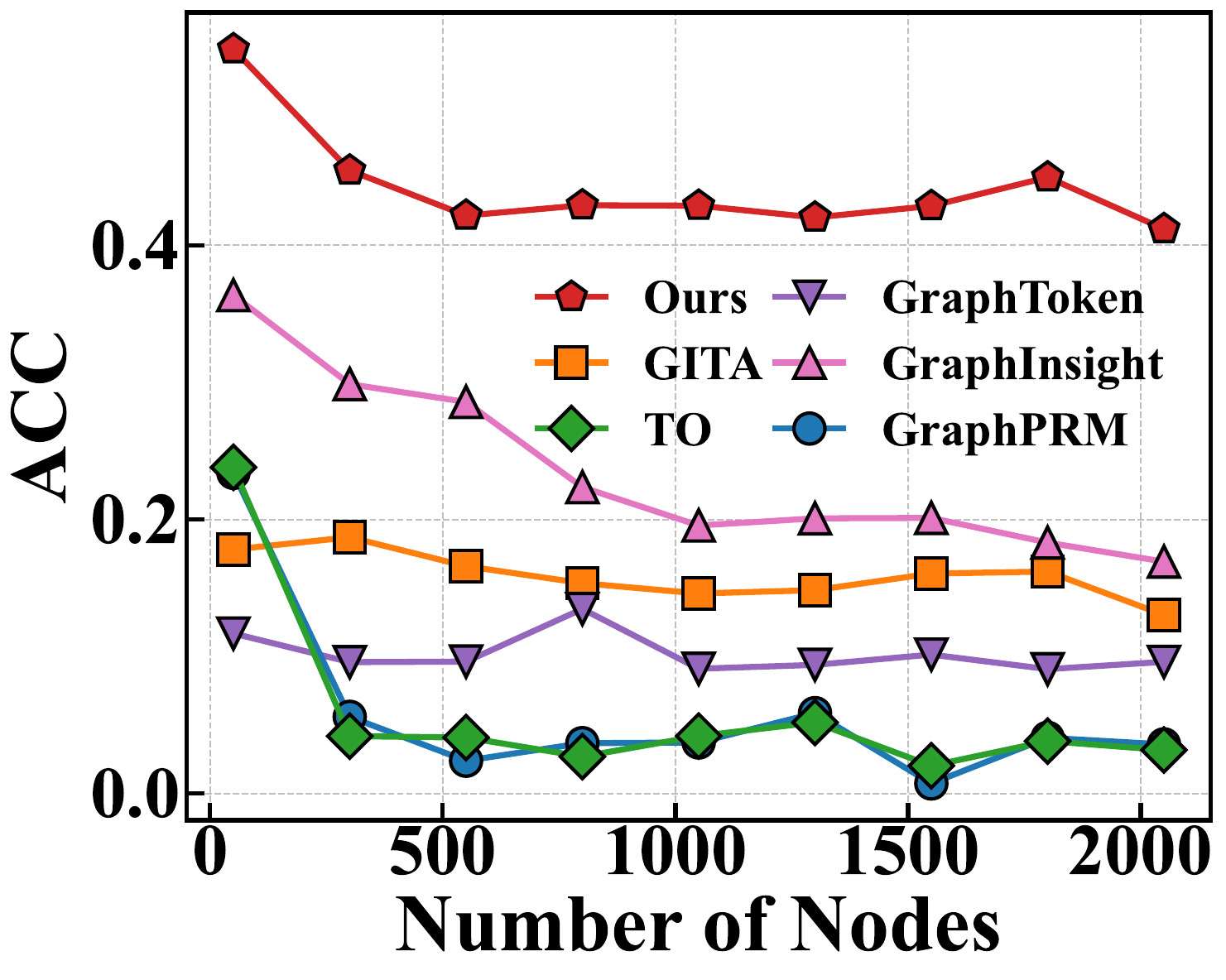}
		\caption{Visual (Gemma-3)}
		\label{fig:gemma_visual}
	\end{subfigure}
	\hfill
	\begin{subfigure}[b]{0.234\textwidth}
		\centering
		\includegraphics[width=\textwidth]{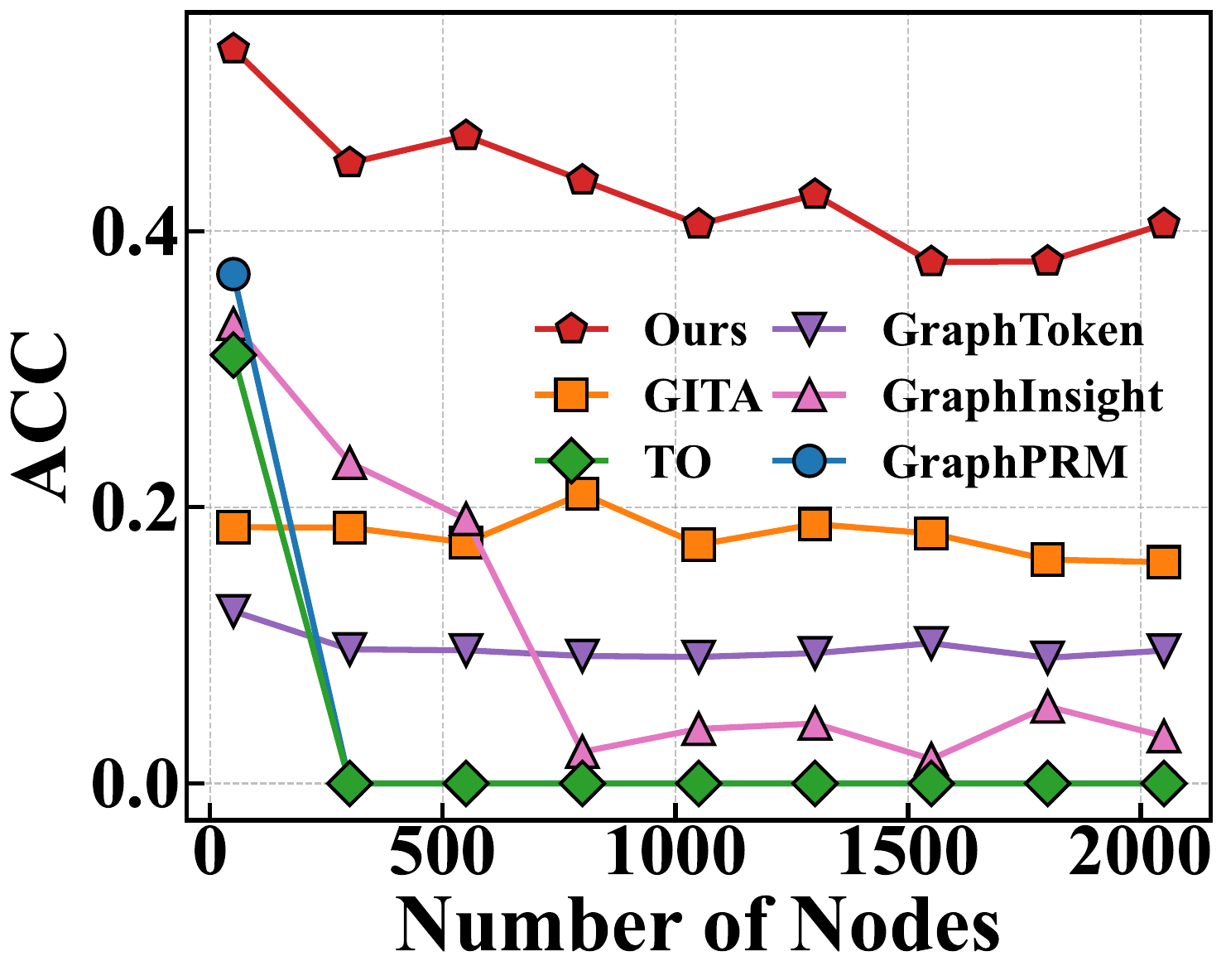}
		\caption{Visual (Qwen2.5-VL)}
		\label{fig:qwen_visual}
	\end{subfigure}

	\begin{subfigure}[b]{0.234\textwidth}
		\centering
		\includegraphics[width=\textwidth]{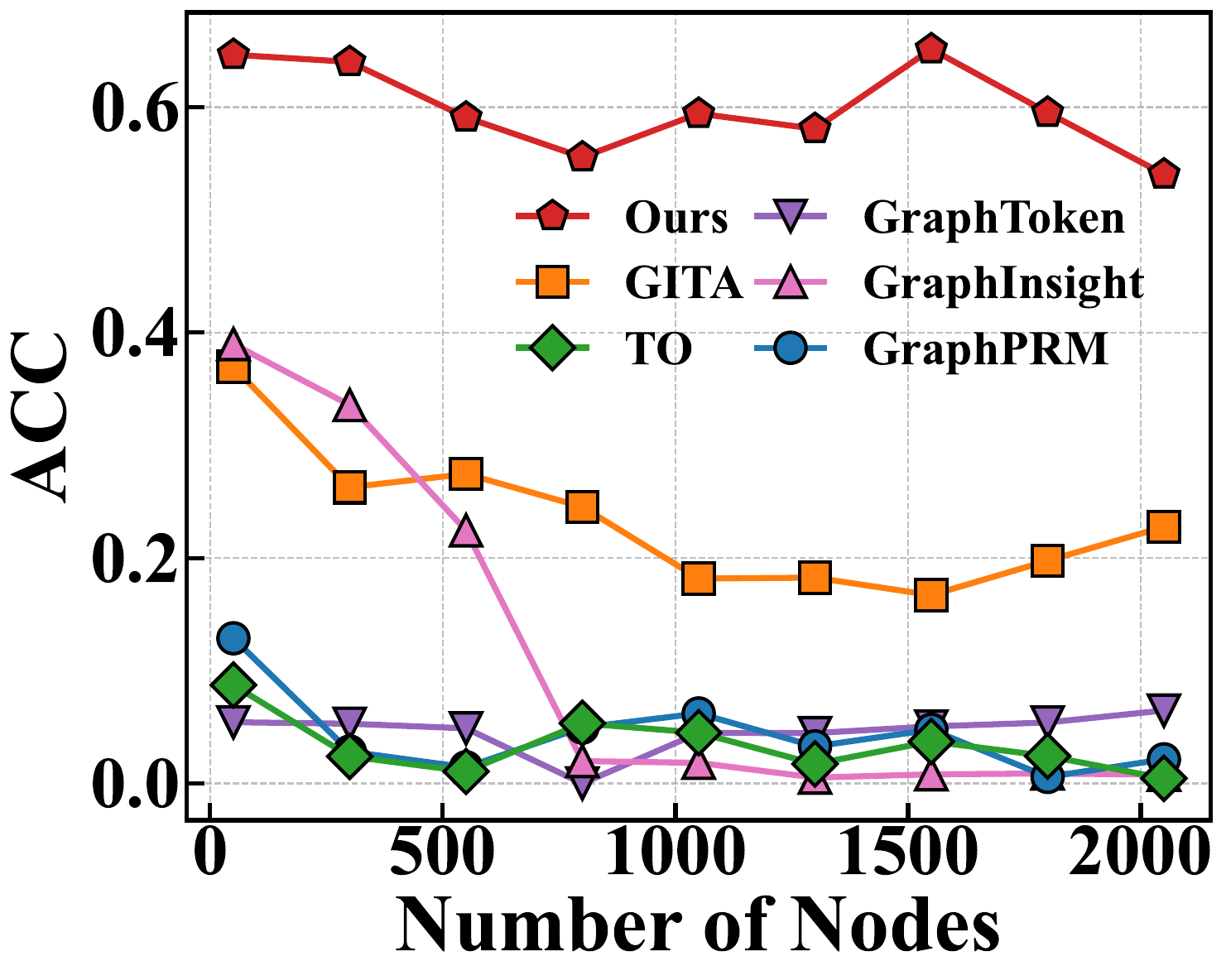}
		\caption{Collab. (Gemma-3)}
		\label{fig:gemma_collab}
	\end{subfigure}
	\hfill
	\begin{subfigure}[b]{0.234\textwidth}
		\centering
		\includegraphics[width=\textwidth]{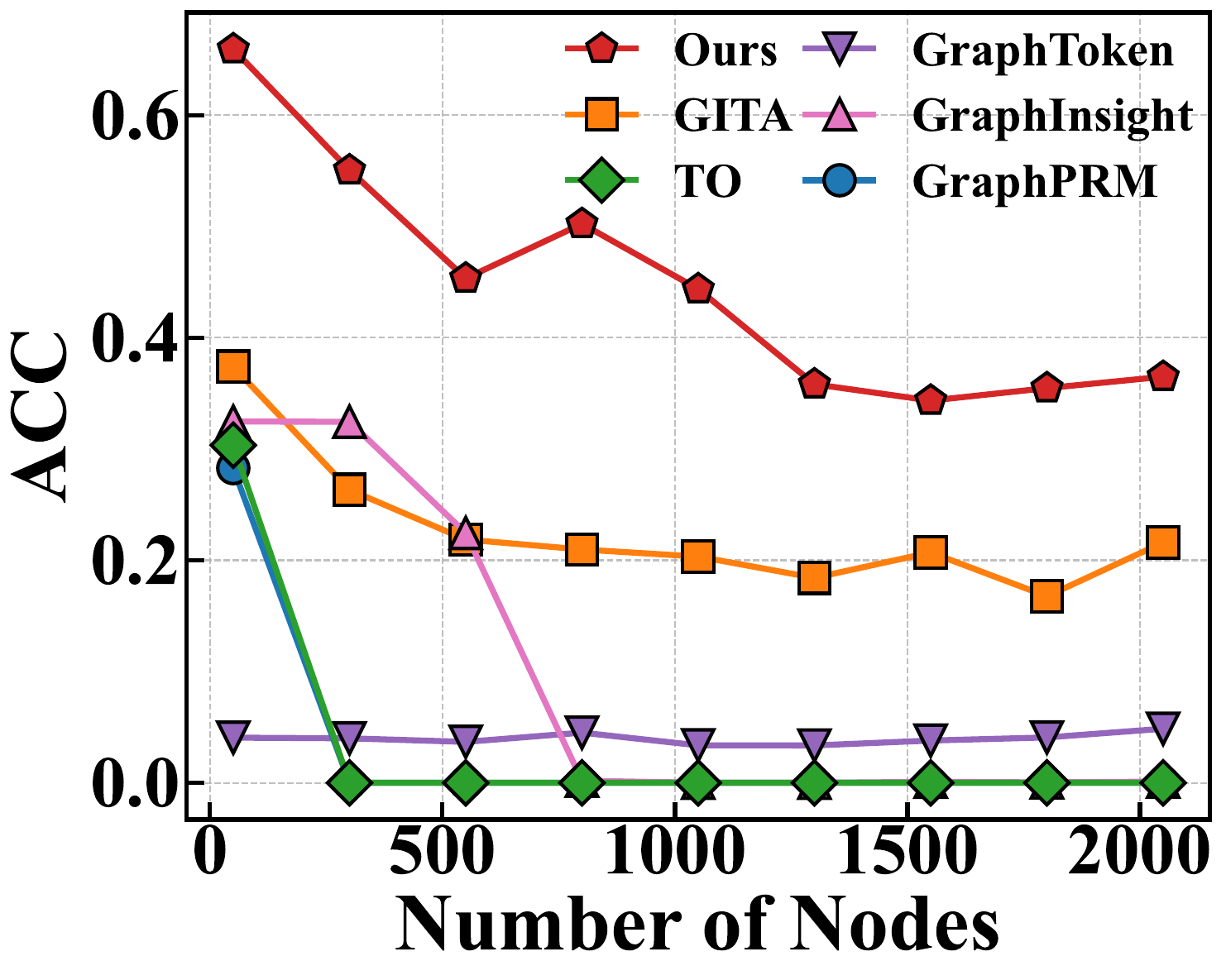}
		\caption{Collab. (Qwen2.5-VL)}
		\label{fig:qwen_collab}
	\end{subfigure}
	
	\caption{Analysis on Graphs with Different $|V|$}
	\label{fig:gemma_qwen_analysis_full}
    \vspace{-4pt}

\end{figure}

\vspace{-3pt}
\subsection{Performance}
\label{sec:performance}
\vspace{-3pt}
As shown in Table~\ref{tab:overall_res}, GraphVista consistently outperforms all baselines on both benchmarks.

\vspace{-3pt}

\paragraph{Text-modality Tasks.}
\label{sec:text_modality}
For text-modality tasks, GraphVista achieves an accuracy of 0.936 on the Grena benchmark, surpassing the strongest baseline, GraphInsight (0.247), by 3.8$\times$. Notably, GraphVista also demonstrates strong generalization on the small-scale GraphSQA benchmark, consistently outperforming baselines across various VLMs. As shown in Figures~\ref{fig:gemma_text} and~\ref{fig:qwen_text}, our method maintains stable performance on Grena up to 2,050 nodes, whereas baseline methods suffer significant degradation (e.g., GraphInsight with Qwen2.5-VL drops from 0.270 to 0.172).

\vspace{-4pt}
\paragraph{Visual and Modality-Collaborative Tasks.}
GraphVista shows robust performance on structural reasoning tasks. On visual-modality tasks, it achieves an average accuracy of 0.443 with Gemma-3, outperforming GITA and GraphToken by 2.4$\times$ and 4.3$\times$, respectively. Similarly, for modality-collaborative tasks, GraphVista reaches 0.602, surpassing the best baseline (GITA, 0.234) by 2.6$\times$. As shown in Figures~\ref{fig:gemma_visual}--\ref{fig:qwen_collab}, GraphVista scales robustly to 2,050 nodes, whereas baselines degrade to near-zero as graph size increases.

\vspace{-4pt}
\subsection{Ablation Study}
\label{exp:as}
\vspace{-1pt}

\paragraph{Reasoning Strategies.}
\vspace{-2pt}
We validate the efficacy of the Visual Graph Thoughts mechanism by comparing it against the standard Chain-of-Thought (CoT) baseline and its text-based variant. As detailed in Table~\ref{tab:ablation_cot}, our approach consistently outperforms these text-dependent methods[cite: 791, 940], notably achieving 1.3$\times$ and 2.6$\times$ accuracy gains on Modality-Collaborative and Visual tasks, respectively. These results confirm that explicit structural grounding is critical for resolving the ambiguities inherent in purely textual reasoning.
\vspace{-3pt}
\paragraph{Impact of DPO.}
Comparing fine-tuned models with frozen baselines (Table~\ref{tab:DPO}) demonstrates that process-level DPO enhances performance across all settings, particularly for complex tasks. Notably, DPO improves Qwen2.5-VL-7B's accuracy on collaborative tasks from 0.3589 to 0.4478. This verifies the effectiveness of aligning model outputs with expert visual reasoning trajectories.

\begin{figure}[t]
	\centering
	% --- 第一行: Gemma-3 ---
	\begin{subfigure}[b]{0.234\textwidth}
		\centering
		\includegraphics[width=\textwidth]{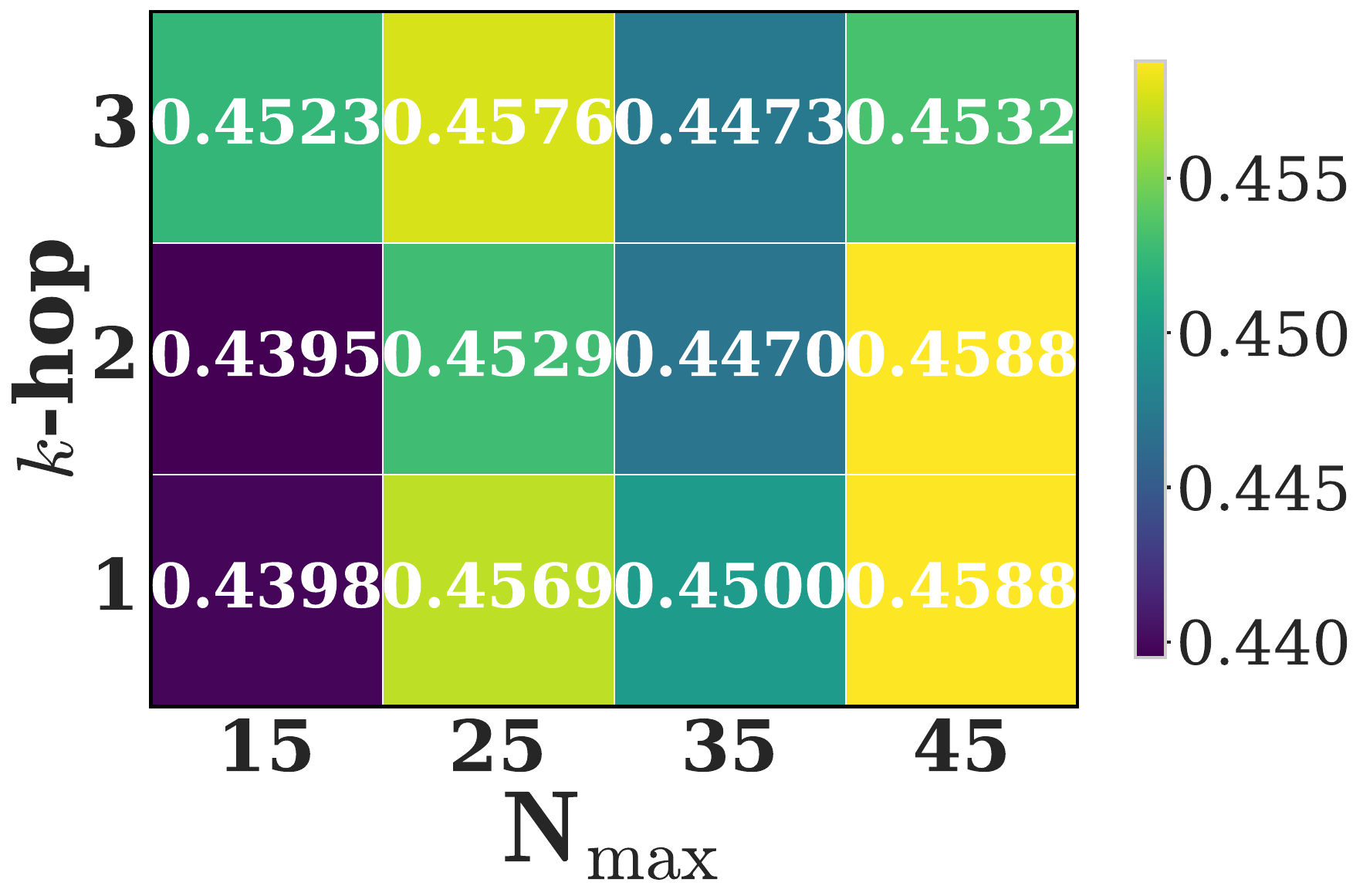}
		\caption{Gemma-3 (Visual)}
		\label{fig:hyp_gemma_visual}
	\end{subfigure}
	\hfill
	\begin{subfigure}[b]{0.234\textwidth}
		\centering
		\includegraphics[width=\textwidth]{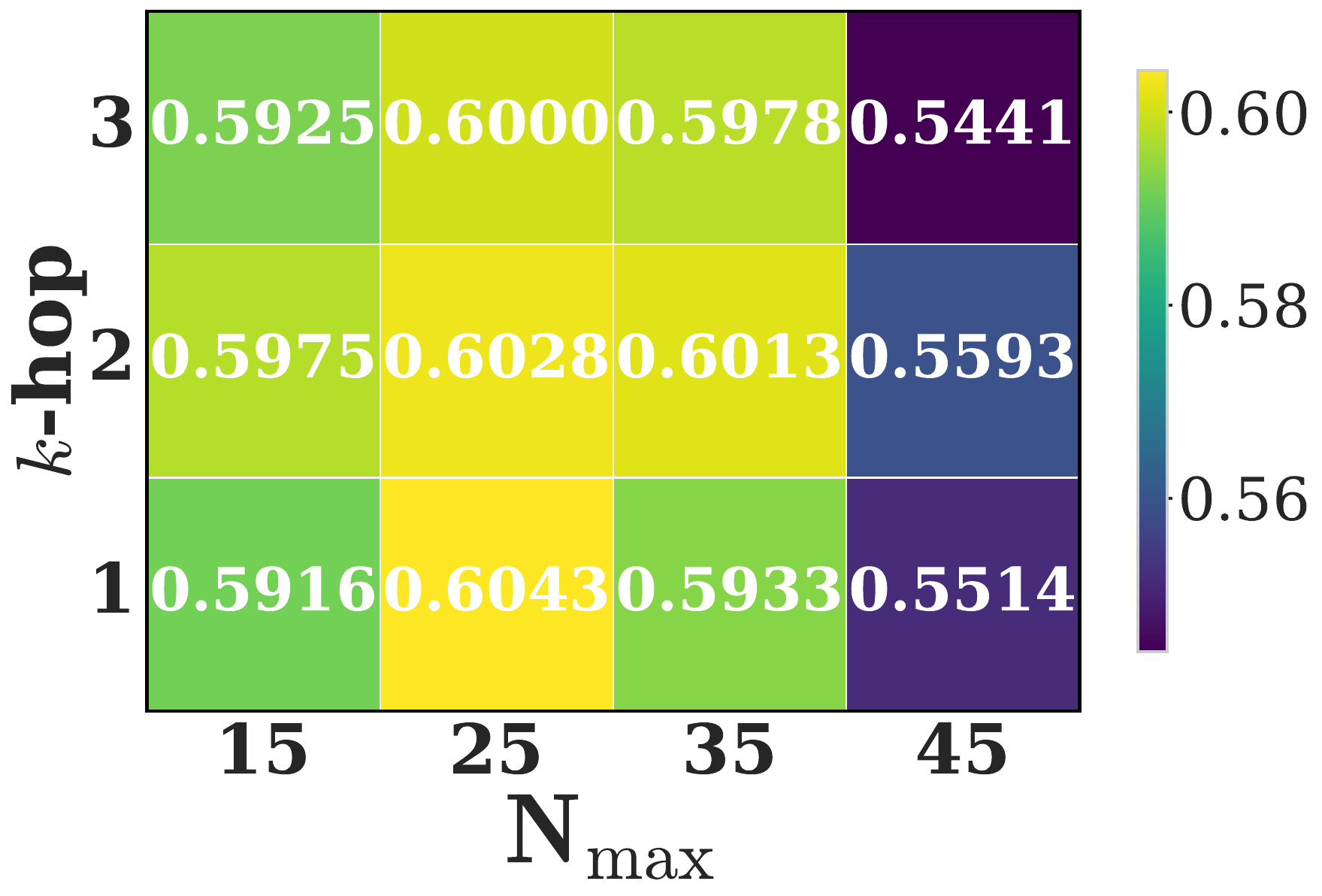}
		\caption{Gemma-3 (Collab.)}
		\label{fig:hyp_gemma_collab}
	\end{subfigure}

	% --- 第二行: Qwen2.5-VL ---
	\begin{subfigure}[b]{0.234\textwidth}
		\centering
		\includegraphics[width=\textwidth]{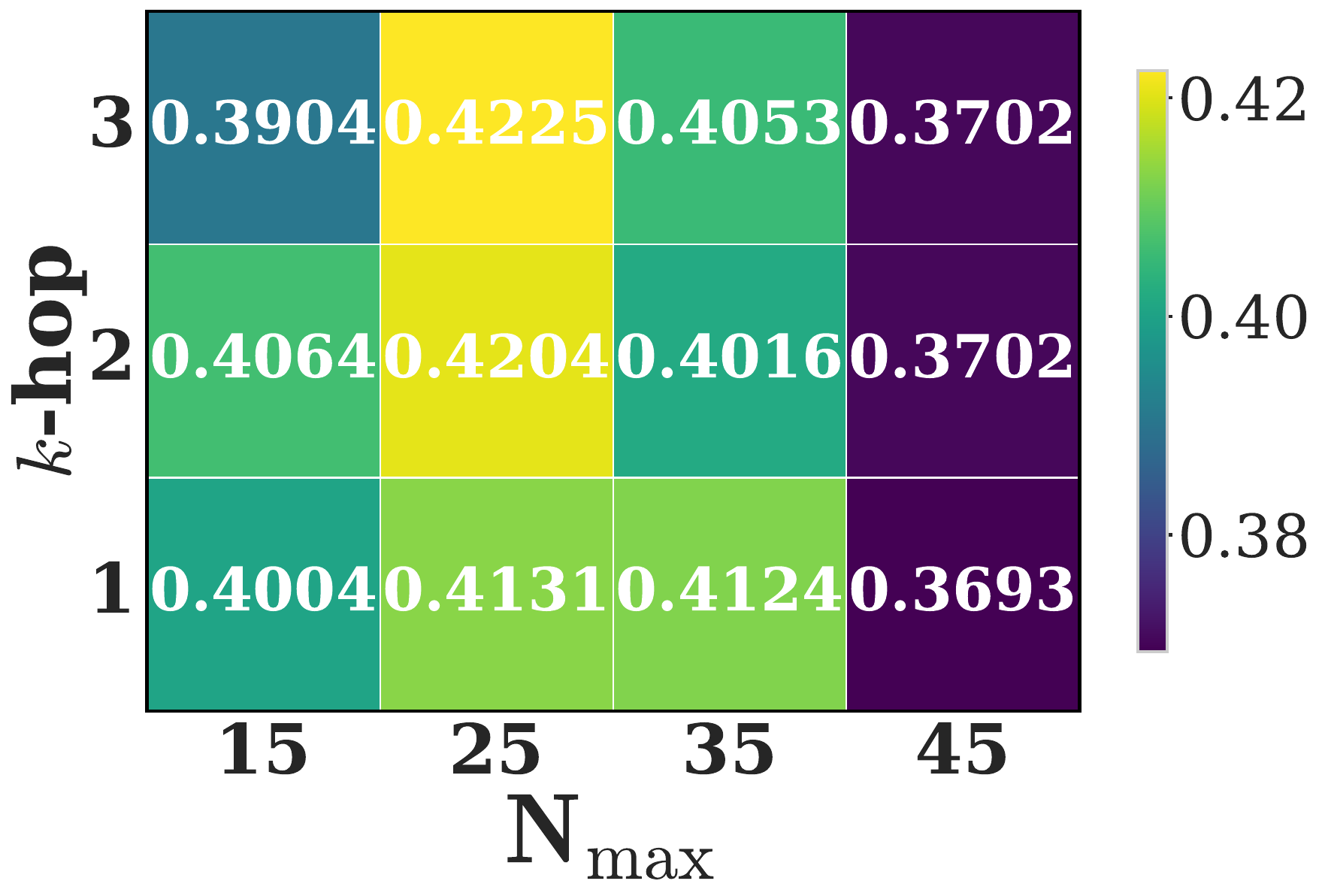}
		\caption{Qwen2.5-VL (Visual)}
		\label{fig:hyp_qwen_visual}
	\end{subfigure}
	\hfill
	\begin{subfigure}[b]{0.234\textwidth}
		\centering
		\includegraphics[width=\textwidth]{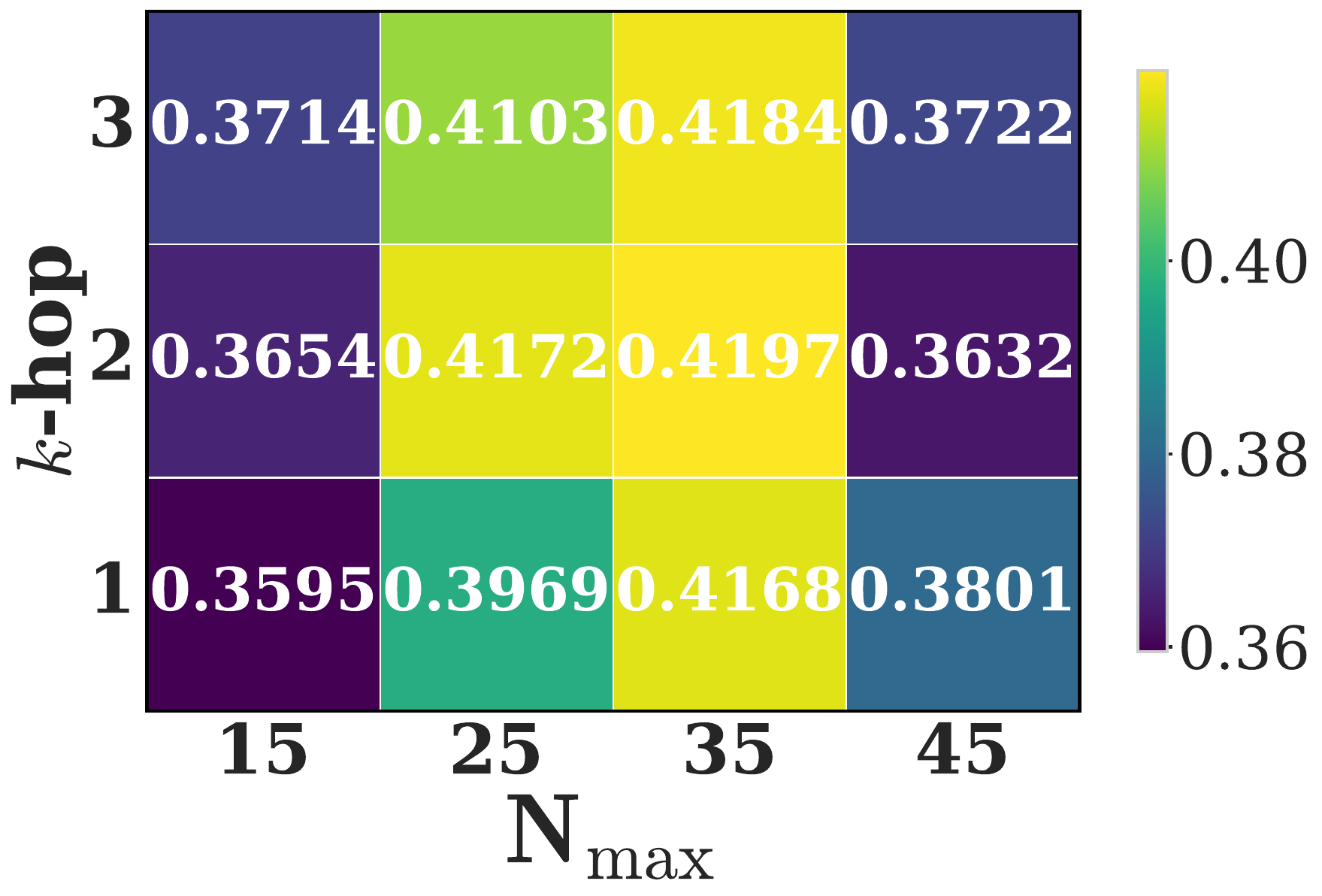}
		\caption{Qwen2.5-VL (Collab.)}
		\label{fig:hyp_qwen_collab}
	\end{subfigure}

	\caption{Hyperparameter analysis on $k$ and $N_{\max}$}
	\label{fig:hyp_subgraph}
	\vspace{-3pt}
\end{figure}

\begin{figure}[t]
\vspace{-10pt}
    \centering
    \includegraphics[width=0.495\textwidth]{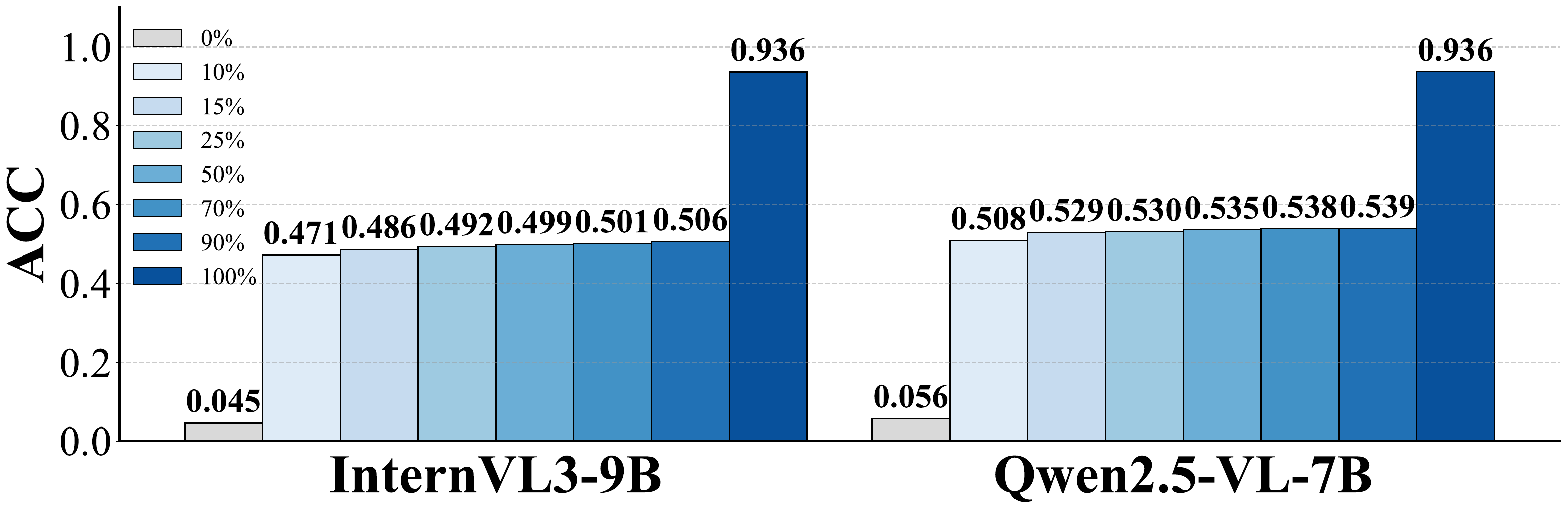}
    \vspace{-15pt}
    \caption{Tier number analysis in $\mathcal{K}$}
    \label{fig:hyp_graphrag}
       \vspace{-5pt}
\end{figure}

\vspace{-4pt}

\subsection{Hyperparameter Analysis}
\label{exp:hp}

\paragraph{Subgraph Extraction.}

\vspace{-3pt}

We analyze the impact of $k$ and $N_{max}$ on performance for visual-modality tasks. As shown in Figure~\ref{fig:hyp_subgraph}, Gemma-3 reaches its highest accuracy with a 2-hop neighborhood, indicating that it can leverage a broader structural context. In contrast, Qwen2.5-VL performs best with a 1-hop neighborhood, suggesting that larger subgraphs may introduce extraneous information. These observations point to the need for adapting subgraph extraction strategies to the characteristics of different VLM architectures.

\vspace{-4pt}

\paragraph{Number of Tiers in $\mathcal{K}$.} We examine the impact of storage granularity in $\mathcal{K}$ across three settings: Tier 1 (Core), Tier 1+2 (Backbone), and Full Graph. As shown in Figure~\ref{fig:hyp_graphrag}, performance positively scales with structural coverage. Tiers 1 and 2 provide efficient baselines, but adding Tier 3 significantly improves performance (e.g., 0.936 on InternVL3-9B) at the cost of higher storage. This underscores a clear trade-off: prioritizing central nodes ensures efficiency, whereas full graph coverage is essential for maximizing accuracy.

\vspace{-4pt}

\section{Conclusion}
\vspace{-3pt}
In this paper, we propose GraphVista, a unified framework for scalable and modality-coordinated understanding of graph structure. For scalability, it employs a hierarchical GraphRAG base to compress task-relevant information. For coordination, a planning agent routes tasks to the optimal modality, leveraging textual retrieval for properties and visual graph thoughts for topological reasoning. Future work will extend GraphVista to complex graphs with labels and semantics.

\newpage
\section*{Limitations}
The proposed GraphVista framework offers a promising approach for scalable, modality-coordinated graph structure understanding, while also presenting several potential extensions that could inspire future research directions.

First, while the framework currently focuses on graph topology, its modular design is intended to facilitate extensions for handling rich semantic information. For instance, a semantic extraction module could be easily incorporated to process node attributes and edge labels. This would broaden the framework's applicability to domains such as knowledge graph reasoning and question answering over heterogeneous networks. Additionally, for the lightweight GraphRAG base, we adopted a general and efficient implementation to ensure the framework's versatility. We note that many advanced optimization strategies, such as sophisticated indexing and retrieval techniques, are orthogonal to our core contributions. A key advantage of GraphVista's modular design is that these specialized modules can be readily integrated into our framework. Therefore, future work could easily incorporate state-of-the-art components to further enhance the performance of our framework, demonstrating its flexibility.
% \clearpage
\bibliography{anthology,custom}

\clearpage
\appendix

\newcommand{\cmark}{\textcolor{green!70!black}{\checkmark}} 
\newcommand{\xmark}{\textcolor{red!80!black}{\ding{55}}}   
\captionsetup{skip=10pt} 

\captionsetup[subfigure]{skip=6pt}

\setlength{\textfloatsep}{20.0pt plus 2.0pt minus 4.0pt}
\setlength{\floatsep}{12.0pt plus 2.0pt minus 2.0pt}
\setlength{\intextsep}{12.0pt plus 2.0pt minus 2.0pt}

\begin{table*}[h!]
	\centering
    \caption{Comparison of Benchmarks for Evaluating graph structure understanding in Large Language Models (LLMs) and Vision-language models (VLMs)}
	\resizebox{\textwidth}{!}{
		\begin{tabular}{lcccc}
			\toprule
			& Max Graph Size & Graph Diversity & Multimodality & Step-Level Visual Reasoning \\
			\midrule
			GraphQA \cite{fatemi2023talk} & 20 & \xmark & \xmark & \xmark \\
			GraphInstruct \cite{luo2024graphinstruct} & 35 & \xmark & \xmark & \xmark \\
			GraphInstruct(Graphwiz) \cite{chen2024graphwiz} & 100 & \xmark & \xmark & \xmark \\
			GraphSQA \cite{cao-etal-2025-graphinsight} & 200 & \cmark & \xmark & \xmark \\
			VisionGraph \cite{li2024visiongraph} & 35 & \xmark & \cmark & \xmark \\
			VGCure \cite{zhu-etal-2025-benchmarking} & 15 & \cmark & \cmark & \xmark \\
			Grena & 2050 & \cmark & \cmark & \cmark \\
			\bottomrule
		\end{tabular}
	}

	\label{tab:benchmark}
\end{table*}

\section{Grena Benchmark}\label{Benchmark}

Existing graph structure understanding benchmarks suffer from three major limitations: 
(i) limited coverage of graph scales and topological structures; 
(ii) lack of task categorization grounded in fundamental graph properties and reasoning modalities; and 
(iii) insufficient support for modeling and evaluating multi-step visual reasoning in graph inference. 

To address these gaps, we propose the \textbf{Grena Benchmark}, a large-scale, multi-task benchmark with step-level visual reasoning capability. It is designed to rigorously evaluate and advance the cognitive and reasoning abilities of VLMs in graph domains. 

The core design of Grena Benchmark is reflected in the following three aspects: 

\begin{table*}[t]
\centering
\caption{Categories of errors used to construct $y_l$.}
\label{tab:error_categories}
\renewcommand{\arraystretch}{1.0} 
\begin{tabular}{@{} l p{0.7\textwidth} @{}}
\toprule

\rowcolor{gray!15}
\multicolumn{2}{l}{\textbf{Textual and Logical Errors}} \\
\cmidrule(r){1-2}
Factual Errors & Incorrect identification of a neighbor set $\mathcal{N}(v)$ or an edge weight $w(e)$. \\
Logical Errors & Faults in algorithmic reasoning (e.g., using edge count instead of weight in a shortest-path task). \\
Computation Errors & Mistakes in numerical operations such as counting. \\
Omitted Steps & Skipping essential reasoning steps. \\
\midrule

\rowcolor{gray!15}
\multicolumn{2}{l}{\textbf{Multimodal Errors}} \\
\cmidrule(r){1-2}
Element Misrecognition & Misidentifying nodes or edges in the visual representation. \\
Visual Neglect & Failure to incorporate visual information, with reasoning based only on text. \\
Text--Visual Inconsistency & Mismatch between textual reasoning and visual grounding. \\
Visualization Misuse & Incorrect or missing calls to visualization functions. \\
\bottomrule
\end{tabular}
\end{table*}

\subsection{Scalability and Structural Diversity}

A primary objective of Grena is to systematically evaluate the scalability graph structure understanding capabilities of VLMs. To this end, the benchmark includes graphs that span a wide range of scales, with node counts from 50 to 2,050. This diversity in size is essential for assessing how model performance varies with increasing graph size and structural complexity.

In addition to scale, Grena incorporates diverse topological structures by including both Erdős–Rényi (ER) random graphs and Barabási–Albert (BA) scale-free networks. While ER graphs provide a baseline for performance on uniform structures, BA graphs feature power-law degree distributions and hubs, closely resembling the topology of many real-world systems. This combination of scales and topologies creates a robust testbed for evaluating model generalization across heterogeneous graph structures.

\subsection{Categorization of Graph Structure Understanding Tasks}

To provide a structured assessment of reasoning abilities across different modalities, Grena Benchmark organizes tasks into three categories: 

\begin{itemize}
    \item \textbf{Text-Modality Tasks.} 
    These tasks focus on basic statistical retrieval and accessing individual node attributes without requiring reasoning over complex topology. Examples include ``What is the total number of nodes?'' or ``What is the degree of node $v_A$?'' They primarily evaluate a model’s ability to interact with the Knowledge Graph ($\mathcal{K}$) for direct information retrieval.

    \item \textbf{Visual-Modality Tasks.} 
    These tasks involve reasoning over local graph structures and topological relationships within a limited subgraph. Examples include shortest-path problems (``Find the shortest path between nodes A and B'') or cycle detection (``Is node A part of any triangle?''). Unlike retrieval tasks, these require the model to utilize visual subgraph representations for efficient structural inference.

    \item \textbf{Modality-Collaborative Tasks.} 
    These tasks require integrating global graph understanding with fine-grained local structural analysis, necessitating a sequence of steps that alternate between modalities. For example, calculating the graph diameter involves identifying peripheral nodes via text modality followed by estimating pairwise distances via visual modality. These tasks evaluate the model's ability to decompose complex queries into sequential sub-tasks.
\end{itemize}

\subsection{Process-Level Supervision for Visual Reasoning}

A key contribution of the Grena Benchmark is its native support for training models on step-level visual reasoning. For tasks that require a chain-of-thought process (e.g., pathfinding), the benchmark provides not only the final answers but also intermediate visual exemplars. These include the original graph and a series of "state graphs" that progressively highlight key nodes, edges, or subgraphs. This design provides dense supervisory signals, making it suitable for both Supervised Fine-Tuning (SFT) and alignment techniques like Direct Preference Optimization (DPO), thereby enabling VLMs to learn explicit reasoning processes.

To facilitate these training paradigms, we construct a high-quality, process-centric dataset where each sample is a triplet $(\mathbf{x}, y_w, y_l)$, representing the input, a preferred reasoning path (Chosen), and a non-preferred one (Rejected):

\begin{itemize}
    \item \textbf{Supervised Fine-Tuning Data (The "Chosen" Path, $y_w$).}
    The "Chosen" paths serve as gold-standard examples for SFT. They are generated based on human-annotated, standardized reasoning templates that align with the logical flow of each task. We use large-scale synthetic graphs (e.g., Erdős–Rényi and Barabási–Albert models), and programmatically validate the correctness and completeness of all reasoning steps using libraries like NetworkX.

    \item \textbf{Direct Preference Optimization Data (The "Rejected" Path, $y_l$).}
    To create preference pairs $(y_w, y_l)$ for DPO, the "Rejected" paths are systematically generated by introducing errors into the "Chosen" paths. These error-construction strategies are adapted from established textual reasoning benchmarks and extended to multimodal contexts. The resulting paths feature common reasoning pitfalls (as detailed in Table~\ref{tab:error_categories}), teaching the model to distinguish between correct and flawed reasoning.
\end{itemize}
\begin{table*}[h!]
\centering
\small
\renewcommand{\arraystretch}{1.15}
\setlength{\tabcolsep}{5pt}
\caption{Taxonomy of tasks in the Grena Benchmark, covering 20 distinct graph structure understanding challenges.}
\label{tab:task_definitions}
\begin{tabular}{p{0.3\linewidth} p{0.65\linewidth}}
\toprule
\textbf{Task Name} & \textbf{Definition and Expected Output} \\
\midrule
Articulation Point Detection & Identify nodes whose removal increases the number of connected components. \textbf{Output}: List of node IDs. \\
Common 3rd-Order Neighbor ID & Find nodes that are exactly 3 hops away from both query nodes $u$ and $v$. \textbf{Output}: Set of node IDs. \\
Connectivity Detection & Determine whether the graph is fully connected. \textbf{Output}: Boolean. \\
Cycle Detection & Determine whether the graph contains any cycles. \textbf{Output}: Boolean. \\
Edge Counting & Return the total number of edges $|E|$. \textbf{Output}: Integer. \\
Edge Existence Checking & Determine if a direct edge exists between a given pair of nodes $(u,v)$. \textbf{Output}: Boolean. \\
Connected Edge Identification & Return all edges incident to a specific query node $v$. \textbf{Output}: Set of edges. \\
Graph Diameter Calculation & Compute the longest shortest path between any two nodes in the graph. \textbf{Output}: Integer. \\
Highest Degree Neighbor ID & For a query node $v$, identify the neighbor with the maximum degree. \textbf{Output}: Node ID. \\
Hub Node Path Finding & Identify a path between two nodes that routes through high-degree hub nodes. \textbf{Output}: Node sequence. \\
Maximum Degree Node ID & Identify the node(s) with the highest degree in the entire graph. \textbf{Output}: Node ID(s). \\
Maximum Clique Detection & Find the size or set of nodes of the largest complete subgraph. \textbf{Output}: Integer (size) or set of nodes. \\
Neighbor Connection Analysis & Count the number of edges within the induced subgraph of a node's neighbors. \textbf{Output}: Integer. \\
Node Counting & Return the total number of nodes $|V|$. \textbf{Output}: Integer. \\
Node Degree Identification & Return the degree of a specific query node $v$. \textbf{Output}: Integer. \\
Planarity Testing & Determine if the graph can be embedded in a plane without edge crossings. \textbf{Output}: Boolean. \\
Shortest Path Finding & Compute the shortest path distance and sequence between nodes $u$ and $v$. \textbf{Output}: Path length and sequence. \\
Star Structure Identification & Determine if a specific subgraph forms a star topology centered at node $v$. \textbf{Output}: Boolean. \\
Third-Order Neighbor ID & Return the set of nodes at exactly distance 3 from a query node $v$. \textbf{Output}: Set of node IDs. \\
Triangle Counting & Count the total number of triangles (3-cycles) in the graph. \textbf{Output}: Integer. \\
\bottomrule
\end{tabular}
\end{table*}

\section{Subgraph Extraction}
\label{appendix_Subgraph_Ext}

To mitigate the scalability challenges of large graphs, we first extract a task-relevant local subgraph $G' = (V', E')$ from $G$ stored in $\mathcal{K}$, based on the key entities $\mathcal{E}$. This process is formalized as:
\begin{equation}
G' = f_{\text{extract}}(G, \mathcal{E})
\label{eq:subgraphextract}
\end{equation}
where the extraction strategy $f_{\text{extract}}$ is determined by the structural characteristics of $\mathcal{E}$ and can be categorized as follows.

\paragraph{Ego-centric}
For problems involving a single entity, where $\mathcal{E} = (v_i, \emptyset)$, we extract its $k$-hop neighborhood subgraph $G'_{v_i, k} = (V', E')$. The node set $V'$ is defined as:
\begin{equation}
V' = \{ v \in V \mid \text{dist}_G(v, v_i) \le k \}
\label{eq:egocentric}
\end{equation}
where $\text{dist}_G(u, v)$ denotes the shortest path distance between nodes $u$ and $v$ in $G$. The edge set $E'$ is induced from $G$ by $V'$.

    \paragraph{Multi-centric} For problems concerning relationships between multiple entities, where $\mathcal{E} = (v_s, v_t)$, we extract paths connecting the source $v_s$ and target $v_t$. We employ Yen's K-shortest paths to obtain $K$ candidate paths $\mathcal{P} = \{P_1, \dots, P_K\}$. We then construct a set $V_{\mathcal{P}} = \bigcup_{j=1}^{K} V(P_j)$ of all nodes on these paths. Finally, a 1-hop neighborhood expansion is performed on all nodes in $V_{\mathcal{P}}$ to form the final subgraph node set $V' = \bigcup_{v \in V_{\mathcal{P}}} \{u \in V \mid \text{dist}_G(u, v) \le 1\}$. The edge set $E'$ is induced by $V'$ from $G$.

\section{Subgraph Pruning and Visualization Strategy}
\label{sec:appendix_c}

To maintain visual clarity and adhere to computational constraints, particularly for large graphs, the size of an extracted subgraph $G'$ is bounded by a maximum node count, $N_{max}$. In instances where the initial extracted subgraph $G'=(V', E')$ exceeds this threshold, a principled pruning procedure is invoked. Furthermore, the remaining nodes are visualized using a task-driven strategy to ensure optimal perception by the VLM.

\subsection{Ego-centric Subgraph Pruning}
This approach, detailed in Algorithm 1, addresses subgraphs generated around a single central node, $v_i$. The core principle is to retain nodes most relevant to this central entity. All nodes except for $v_i$ are considered candidates for removal and are sorted into a priority list for pruning based on the hierarchical key $(-d(v, v_i), t(v), c(v))$. This multi-level sorting criterion prioritizes the removal of nodes based on the following order:

\begin{enumerate}
    \item \textbf{Distance:} Nodes farthest from the central node $v_i$ (descending distance $d$) are removed first.
    \item \textbf{Structural Tier:} Among nodes at the same distance, those in lower-importance tiers within the Hierarchical GraphRAG Base $\mathcal{K}$ (ascending tier $t(v)$) are prioritized for removal.
    \item \textbf{Centrality:} For nodes with the same distance and tier, those with a lower centrality score (ascending centrality $c(v)$) are removed.
\end{enumerate}
By systematically applying these criteria, the algorithm ensures that the pruned subgraph retains the most structurally-important neighborhood around the central node.

\subsection{Multi-centric Subgraph Pruning}
This strategy, presented in Algorithm 2, is designed for subgraphs constructed around a set of K-shortest paths, $V_{\mathcal{P}}$, connecting multiple entities. The primary objective is to preserve the integrity of these critical paths. Therefore, all nodes lying on these paths ($V_{\mathcal{P}}$) are explicitly protected from pruning. Candidate nodes for removal are those not on the primary paths, and their sorting priority is determined by the key $(t(v), c(v))$, which prioritizes removal based on:

\begin{enumerate}
    \item \textbf{Structural Tier:} Nodes in lower-importance tiers (ascending tier $t(v)$).
    \item \textbf{Centrality:} Among nodes in the same tier, those with a lower centrality score (ascending centrality $c(v)$).
\end{enumerate}
This approach effectively reduces the subgraph's complexity while guaranteeing that the core relational structure between the key entities remains intact.

\subsection{Task-Driven Subgraph Visualization Strategy}
The visual representation of the extracted subgraph $G'$ plays a critical role in the VLM's reasoning capability. We employ a \textbf{task-driven visualization strategy} where the graph layout algorithm is dynamically selected based on the task type $T$:

\begin{itemize}
    \item \textbf{Hierarchical Layouts:} For sequential reasoning tasks such as \textit{Shortest Path} or \textit{Cycle Detection}, we utilize hierarchical layouts (e.g., Sugiyama algorithm). This layout emphasizes directionality and flow, enabling the VLM to visually trace paths layer-by-layer effectively.
    \item \textbf{Force-Directed Layouts:} For structural analysis tasks such as \textit{Community Detection} or \textit{Clustering Coefficient}, we employ force-directed algorithms (e.g., Fruchterman-Reingold). This approach naturally clusters densely connected nodes, making community structures visually distinct.
\end{itemize}

\paragraph{Reproducibility and Clarity.} To ensure strict reproducibility of the visual inputs, we fix the random seeds for all layout generation processes. Furthermore, to handle potential visual clutter in high-density subgraphs, we apply a \textbf{visual anti-overlap mechanism}. Following the pruning process, we dynamically adjust node repelling forces in the layout engine to ensure that no two nodes overlap and that edge crossings are minimized. This provides a clear ``what-you-see-is-what-you-get'' input for the Visual Graph Thoughts agent.

\SetAlFnt{\small}
\SetAlCapFnt{\small}
\SetAlCapNameFnt{\small}
\setlength{\algomargin}{1em}
\setlength{\textfloatsep}{5pt}
\setlength{\floatsep}{5pt}
\SetKwComment{Comment}{/* }{ */}
\DontPrintSemicolon

\begin{algorithm}[t]
    \caption{Ego-centric Subgraph Pruning}
    \label{alg:ego_pruning}
    \SetKwInOut{Input}{Input}
    \SetKwInOut{Output}{Output}
    \Input{
        $G'=(V', E')$: Initial $k$-hop neighborhood subgraph.\\
        $v_i$: Central node for extraction.\\
        $N_{\max}$: Maximum allowed number of nodes.\\
        $\mathcal{K}$: Hierarchical GraphRAG Base.
    }
    \Output{$G'_{\text{pruned}}$: Pruned subgraph.}
    \BlankLine
    
    \lIf{$|V'| \leq N_{\max}$}{\Return{$G'$}}
    
    $V_{\text{cand}} \gets V' \setminus \{v_i\}$\;

    \tcp*[l]{Sort priority for removal: Farthest, Tier 3 (Peripheral), Lowest Centrality}
    Sort $V_{\text{cand}}$ into list $L$ by key $(-d(v, v_i), -t(v), c(v))$\;
    
    $n_{\text{prune}} \gets |V'| - N_{\max}$\;
    $V_{\text{prune}} \gets L[1:n_{\text{prune}}]$\;
    $V_{\text{kept}} \gets V' \setminus V_{\text{prune}}$\;
    $G'_{\text{pruned}} \gets$ subgraph of $G'$ induced by $V_{\text{kept}}$\;
    \Return{$G'_{\text{pruned}}$}\;
\end{algorithm}

\begin{algorithm}[t]
    \caption{Multi-centric Subgraph Pruning}
    \label{alg:multi_pruning}
    \SetKwInOut{Input}{Input}
    \SetKwInOut{Output}{Output}
    \Input{
        $G'=(V', E')$: Subgraph expanded from $K$-shortest paths.\\
        $V_{\mathcal{P}}$: Nodes on the $K$-shortest paths.\\
        $N_{\max}$: Maximum allowed number of nodes.\\
        $\mathcal{K}$: Hierarchical GraphRAG Base.
    }
    \Output{$G'_{\text{pruned}}$: Pruned subgraph.}
    \BlankLine
    
    \lIf{$|V'| \leq N_{\max}$}{\Return{$G'$}}
    
    $V_{\text{cand}} \gets V' \setminus V_{\mathcal{P}}$\;
    \tcp*[l]{Sort by priority: tier $\uparrow$, centrality $\uparrow$}
    Sort $V_{\text{cand}}$ into list $L$ by key $(t(v), c(v))$\;
    
    $n_{\text{prune}} \gets |V'| - N_{\max}$\;
    $V_{\text{prune}} \gets L[1:n_{\text{prune}}]$\;
    $V_{\text{kept}} \gets V' \setminus V_{\text{prune}}$\;
    $G'_{\text{pruned}} \gets$ subgraph of $G'$ induced by $V_{\text{kept}}$\;
    \Return{$G'_{\text{pruned}}$}\;
\end{algorithm}

\begin{table}[h!]
\centering
\caption{Key Hyperparameters and Framework Configurations.}
\label{tab:hyperparameters}
\resizebox{\columnwidth}{!}{%
\begin{tabular}{@{}llc@{}}
\toprule
\textbf{Category} & \textbf{Parameter} & \textbf{Value} \\
\midrule
\multicolumn{3}{l}{\textit{\textbf{Model \& Inference Configuration}}} \\
Tensor Parallel & \texttt{TENSOR\_PARALLEL\_SIZE} & 1 \\
GPU Memory Util. & \texttt{GPU\_MEMORY\_UTILIZATION} & 0.4 \\
Max Model Length & \texttt{MAX\_MODEL\_LEN} & 4096 \\
\midrule
\multicolumn{3}{l}{\textit{\textbf{Generation Sampling Parameters}}} \\
Temperature & \texttt{TEMPERATURE} & 0.01 \\
Max Tokens & \texttt{MAX\_TOKENS} & 2048 \\
Top P & \texttt{TOP\_P} & 0.9 \\
\midrule

\multicolumn{3}{l}{\textit{\textbf{Subgraph Extraction}}} \\
Max Subgraph Nodes & \texttt{max\_nodes} & 25 \\
Max Hops & \texttt{max\_hops} & 2 \\
\bottomrule
\end{tabular}%
}
\end{table}
\section{Notations}\label{app-notations}
This section summarizes all notations used throughout this paper, as in Table~\ref{tab-symbols}.

\begin{table}[t]
    \centering
    \small
    \caption{Summary of notations used in this paper.}
    \label{tab-symbols}

    \begin{tabularx}{\columnwidth}{@{}l X@{}}
        \toprule
        \textbf{Notation} & \textbf{Description} \\
        \midrule
        $G=(V,E)$ & Graph with node set $V$ and edge set $E$. \\
        $\mathcal{K}$ & The Hierarchical GraphRAG Base. \\
        $Q, T, \mathcal{E}$ & NL question, its task type, and key entities. \\
        $f_{\text{parse}}$ & Semantic parsing function for questions. \\
        $\mathcal{N}_{k}(v)$ & $k$-hop neighborhood of node $v$. \\
        $|V|, N_{\text{max}}$ & Total nodes in graph; max nodes in a subgraph. \\
        $K_1\%, K_2\%$ & Node percentage in Tier 1 (Core) and Tier 2 (Backbone). \\
        \addlinespace
        $G'=(V',E')$ & A subgraph extracted from the original graph $G$. \\
        $f_{\text{viz}}, G_{\text{image}}$ & Subgraph visualization function and its visual output. \\
        $\mathcal{M}_{\text{VRA}}$ & The Visual Reasoning Agent. \\
        $\Pi, S_t, H_t$ & Reasoning plan, state at step $t$, and history up to $t$. \\
        $(o_t, a_t)$ & Intermediate output and action at step $t$. \\
        $C_{\text{retrieved}}$ & Retrieved context from the GraphRAG Base $\mathcal{K}$. \\
        $f_{\text{retrieve}}$ & Retrieval function for querying $\mathcal{K}$. \\
        \addlinespace
        $\mathcal{D}, (x, y_w, y_l)$ & Preference dataset and sample (input, chosen, rejected). \\
        $\pi_{\theta}, \pi_{\text{ref}}$ & DPO policy and fixed reference models. \\
        $r_{\theta}(y|x)$ & Ratio of policy probabilities $\pi_{\theta}(y|x)/\pi_{\text{ref}}(y|x)$. \\
        $\beta, \mathcal{L}_{\text{DPO}}$ & DPO regularization strength and its loss function. \\
        \addlinespace
        $f_{\text{extract}}$ & The subgraph extraction function. \\
        $dist_G(u,v)$ & Shortest path distance between nodes $u$ and $v$. \\
        $V_{\mathcal{P}}$ & Set of nodes on the K-shortest paths. \\
        $d(v, v_i), t(v), c(v)$ & Node's distance to center $v_i$, structural tier, and centrality. \\
        \bottomrule
    \end{tabularx}
\end{table}

\section{Experimental Settings}
\label{sec:experimental_settings}

Our experiments are conducted using 8 NVIDIA H20 GPUs. The framework is implemented in Python 3.11 with CUDA Version 12.6, leveraging the \texttt{vLLM} library for efficient inference of the VLMs.

\subsection{Experimental Settings and Hyperparameters}
Our experiments are conducted using 8 NVIDIA H20 GPUs. The framework is implemented in Python 3.11 with CUDA Version 12.6, leveraging the VLLM library for efficient inference of the VLMs. Key hyperparameters and framework configurations are detailed in Table \ref{tab:hyperparameters}.

\subsection{Process-Level DPO Training Details}
We fine-tuned $\mathcal{M}_{VRA}$ (instantiated with Gemma-3-12B and Qwen2.5-VL-7B) using the curated preference dataset $\mathcal{D}$. The DPO training was performed with a regularization parameter $\beta = 0.1$. We utilized a learning rate of $5e^{-7}$ with a cosine decay scheduler and an effective batch size of 16 (achieved via gradient accumulation). To ensure memory efficiency during training, we employed LoRA (Low-Rank Adaptation) with rank $r=64$ and $\alpha=128$.

The preference dataset $\mathcal{D}$ consists of \textbf{10,221 samples}, specifically constructed to cover diverse reasoning failure modes. The distribution of error types in the rejected trajectories ($y_l$) is summarized in Table~\ref{tab:error-distribution}.

\begin{table}[t]
\centering
\resizebox{\columnwidth}{!}{
\begin{tabular}{lcc}
\toprule
\textbf{Error Type} & \textbf{Proportion} & \textbf{Count} \\
\midrule
Multimodal Errors            & 35\% & 3,577 \\
Textual and Logical Errors   & 65\% & 6,644 \\
\bottomrule
\end{tabular}
}
\caption{Distribution of error types in the rejected trajectories of the preference dataset.}
\label{tab:error-distribution}
\end{table}

This balanced distribution ensures the model learns to robustly ground its reasoning in visual data while maintaining logical coherence.

\begin{figure*}[htbp]
    \centering
    \begin{subfigure}{0.23\linewidth}
        \includegraphics[width=\linewidth]{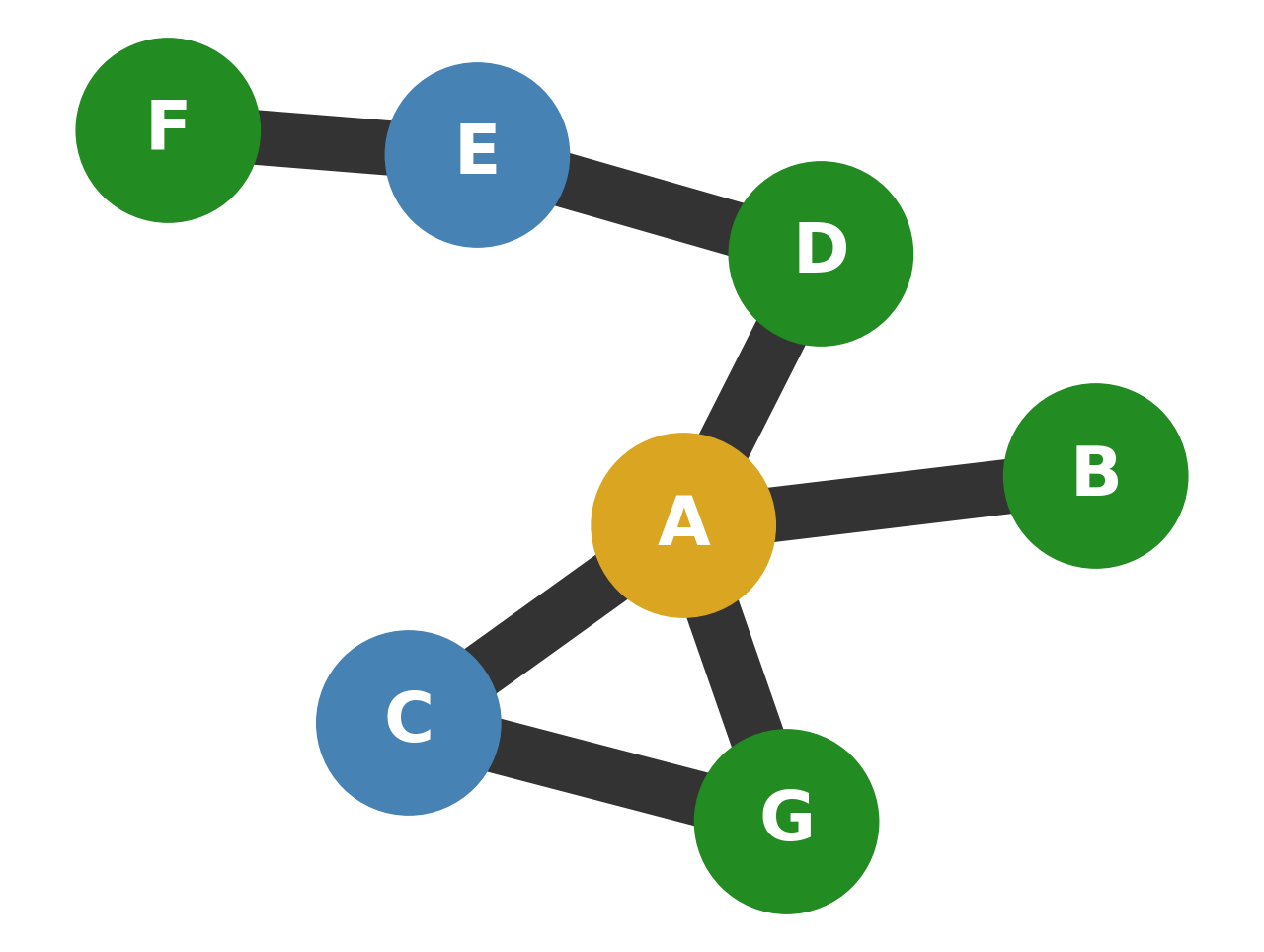}
        \caption{Original graph}
        \label{fig:triangle_a_opt}
    \end{subfigure}
    \hfill
    \begin{subfigure}{0.23\linewidth}
        \includegraphics[width=\linewidth]{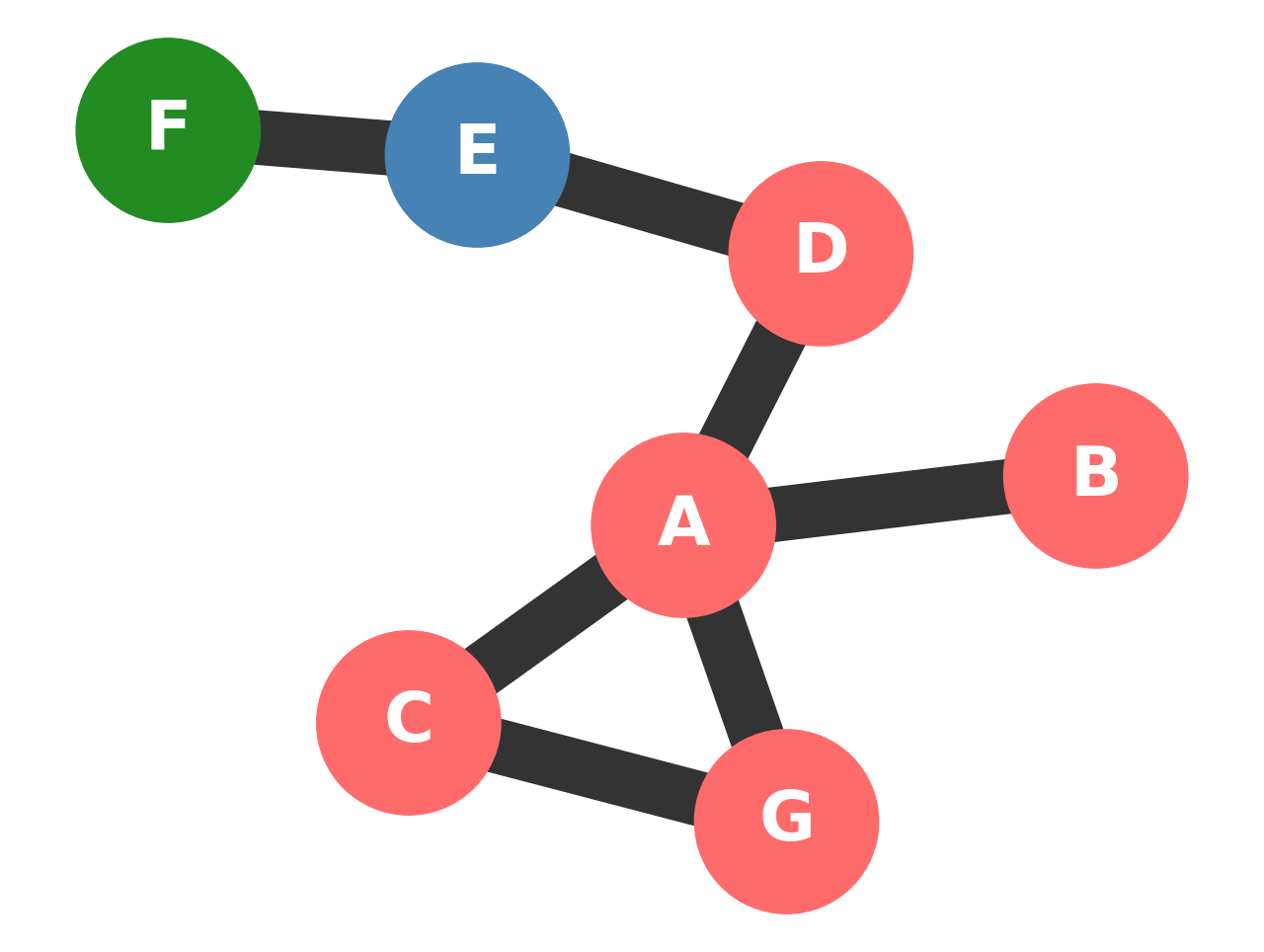}
        \caption{Identify neighbors}
        \label{fig:triangle_b_opt}
    \end{subfigure}
    \hfill
    \begin{subfigure}{0.23\linewidth}
        \includegraphics[width=\linewidth]{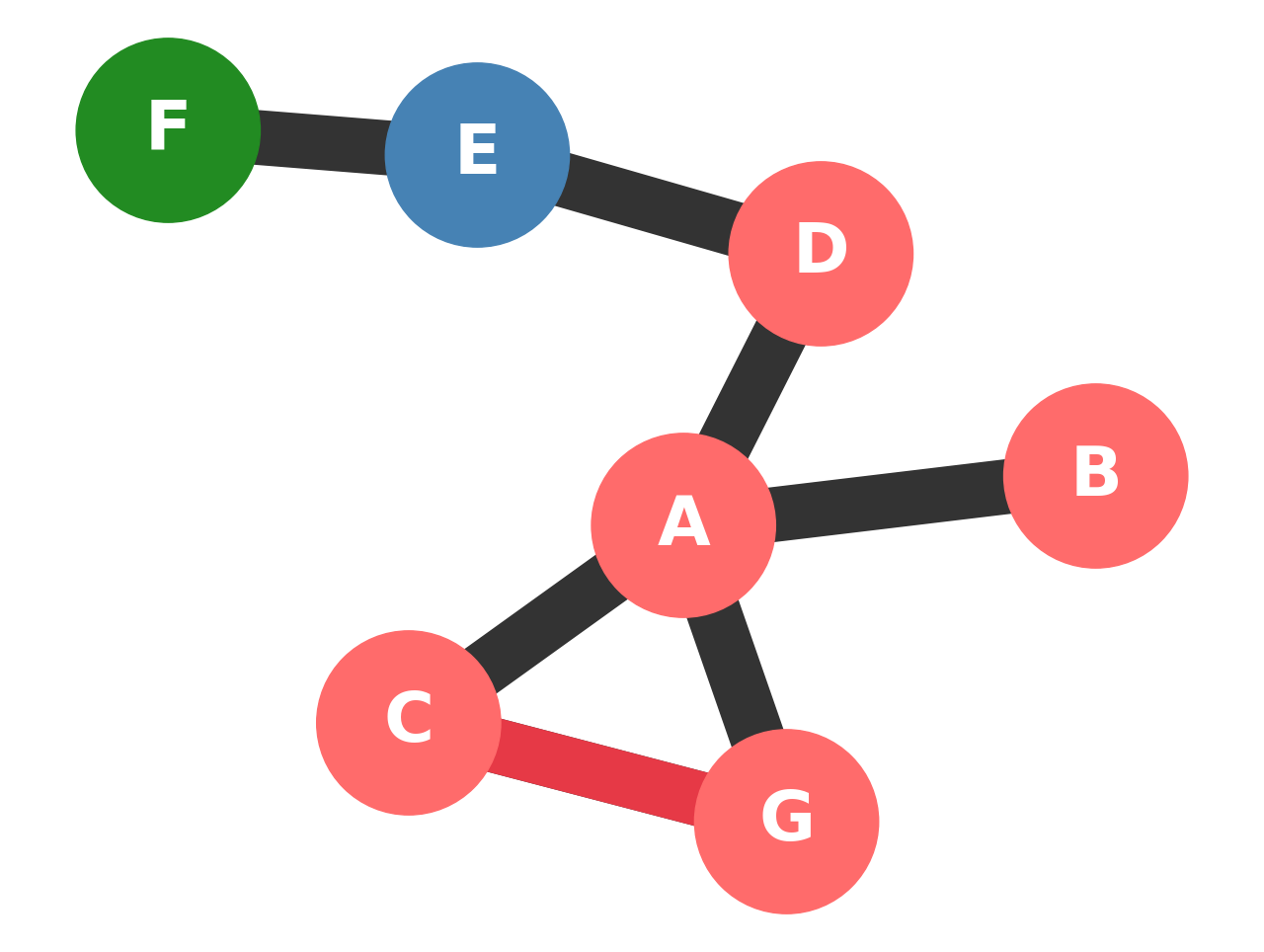}
        \caption{Check edges}
        \label{fig:triangle_c_opt}
    \end{subfigure}
    \hfill
    \begin{subfigure}{0.23\linewidth}
        \includegraphics[width=\linewidth]{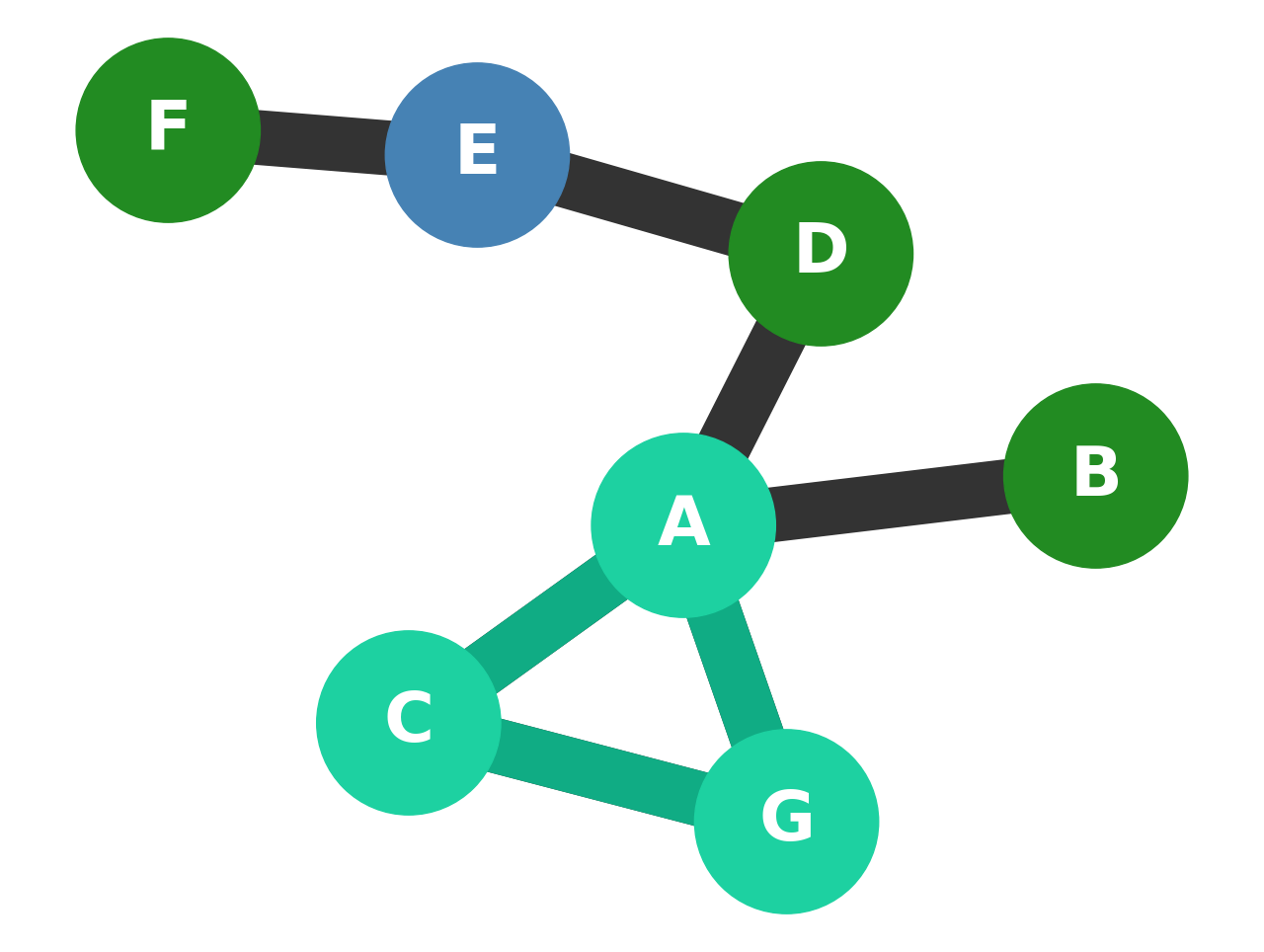}
        \caption{Triangle confirmed}
        \label{fig:triangle_d_opt}
    \end{subfigure}
    \caption{Visualizing the process of \textbf{Triangle Detection} for node A. The algorithm identifies node A's neighbors and then checks for edges between them to confirm a 3-cycle.}
    \label{fig:triangle_detection_optimized}
\end{figure*}

\begin{figure*}[htbp]
    \centering
    % --- Top Row: Three images, evenly spaced ---
    \begin{subfigure}{0.3\linewidth}
        \includegraphics[width=\linewidth]{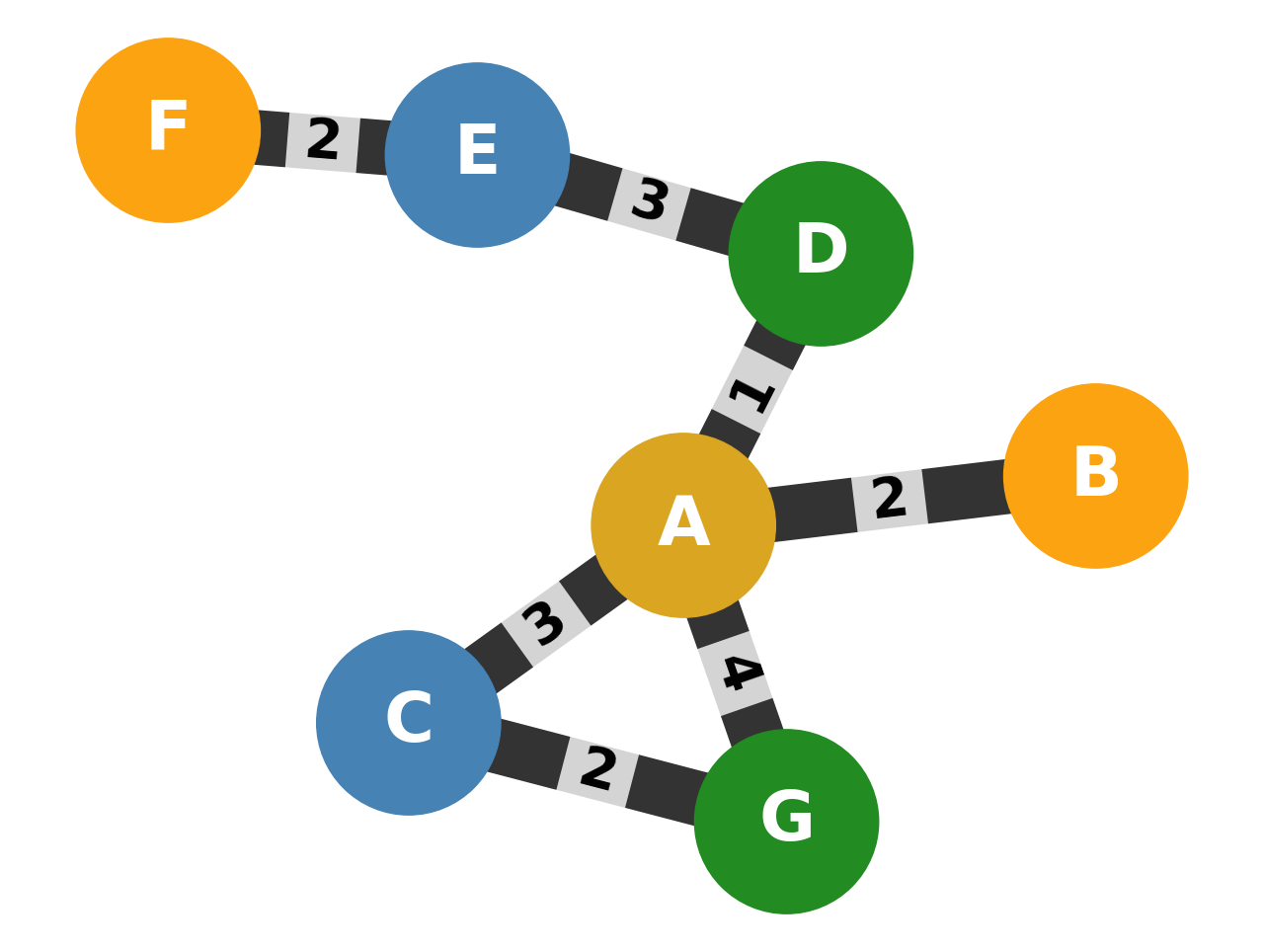}
        \caption{Identify start (B) and end (F)}
        \label{fig:path_a_opt}
    \end{subfigure}
    \hfill
    \begin{subfigure}{0.3\linewidth}
        \includegraphics[width=\linewidth]{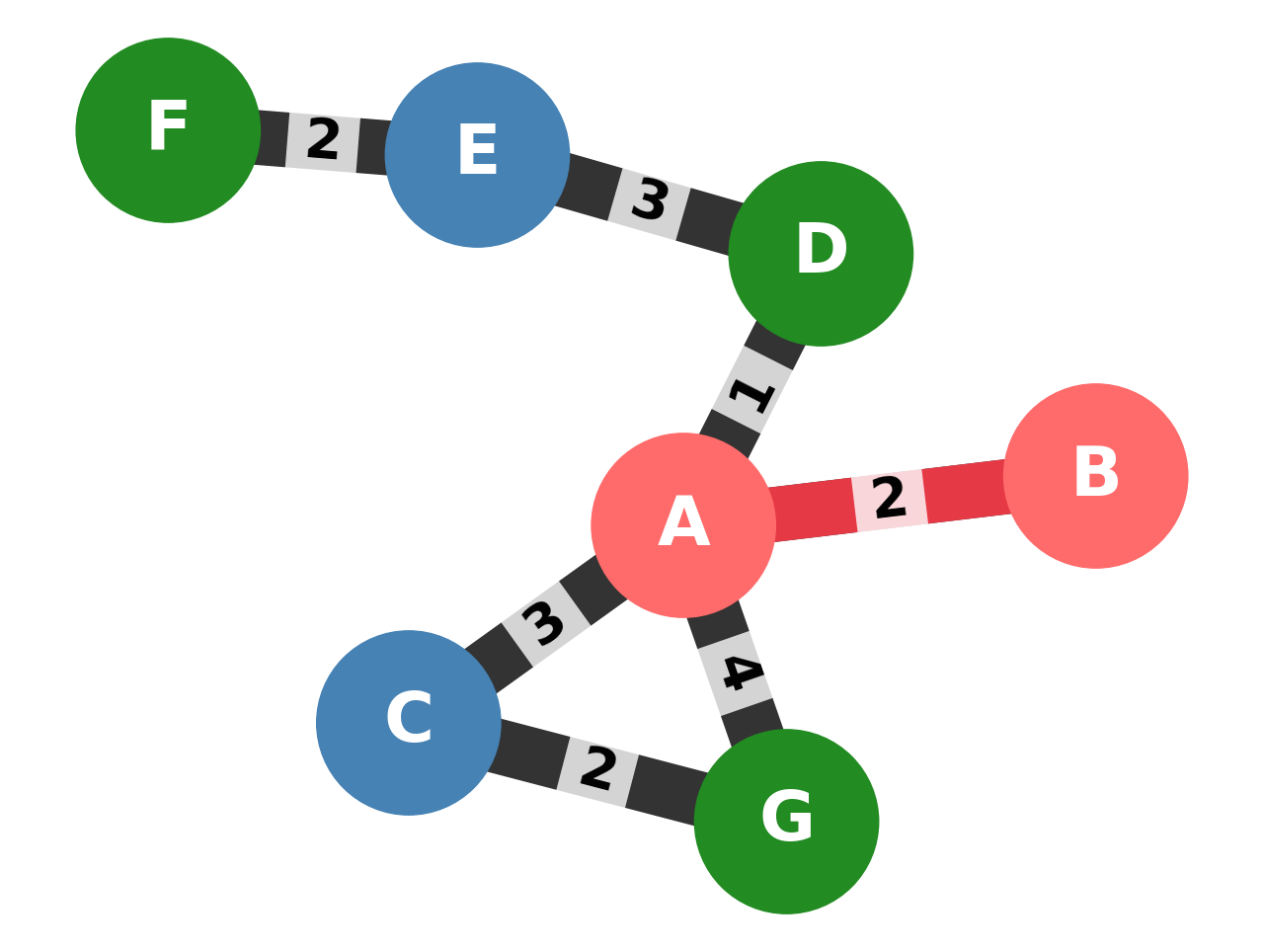}
        \caption{Explore from node B}
        \label{fig:path_b_opt}
    \end{subfigure}
    \hfill
    \begin{subfigure}{0.3\linewidth}
        \includegraphics[width=\linewidth]{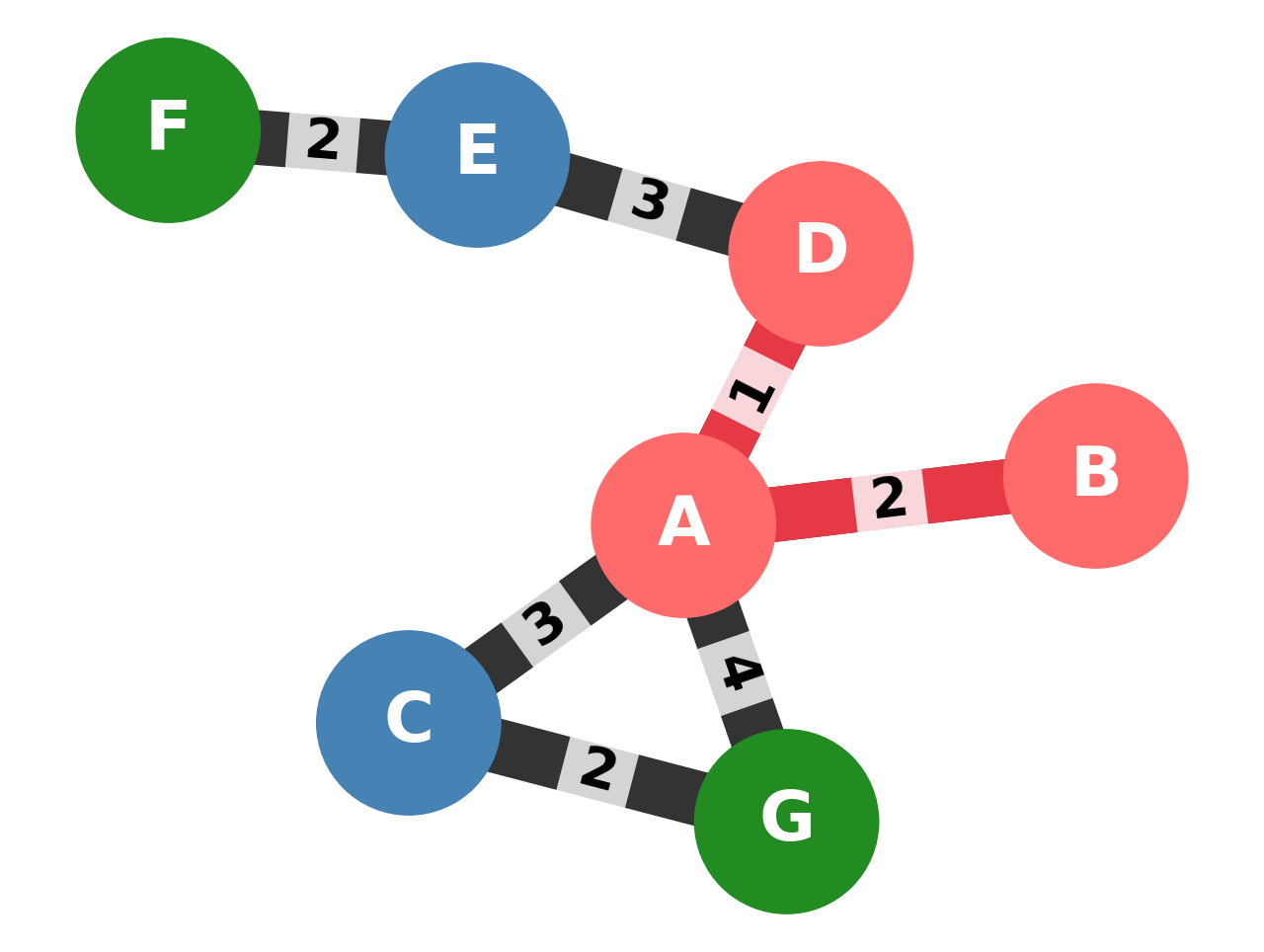}
        \caption{Path extended to node D}
        \label{fig:path_c_opt}
    \end{subfigure}
    
    \vspace{1em} 

    \begin{subfigure}{0.3\linewidth}
        \includegraphics[width=\linewidth]{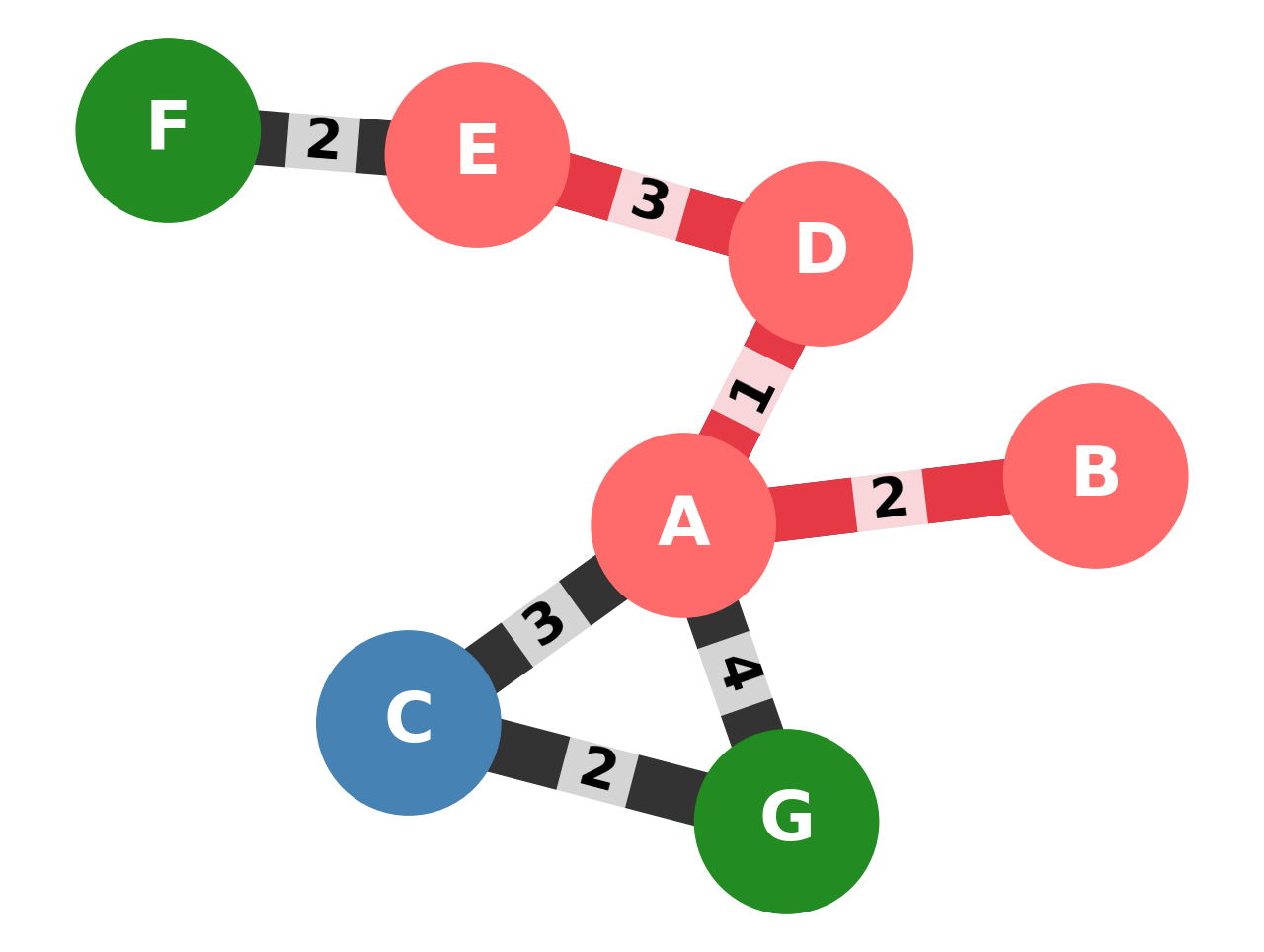}
        \caption{Path extended to node E}
        \label{fig:path_d_opt}
    \end{subfigure}
    \hspace{2em}
    \begin{subfigure}{0.3\linewidth}
        \includegraphics[width=\linewidth]{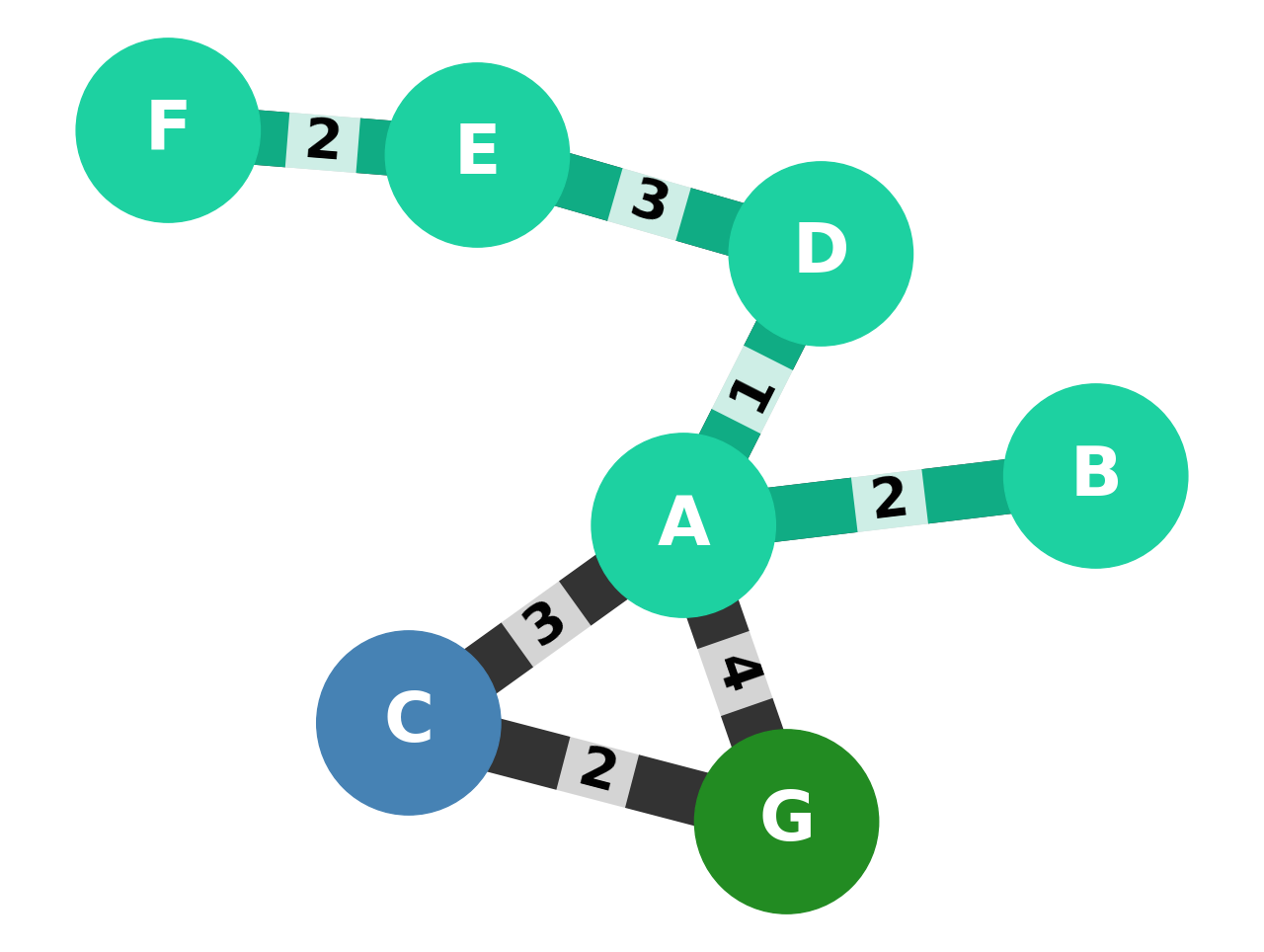}
        \caption{Shortest path found}
        \label{fig:path_e_opt}
    \end{subfigure}
    
    \caption{Visualizing a breadth-first search for the \textbf{Shortest Path} from node B to F. The algorithm explores neighbors layer-by-layer until reaching the destination.}
    \label{fig:shortest_path_optimized}
\end{figure*}

\begin{figure*}[htbp]
    \centering
    \begin{subfigure}{0.3\linewidth}
        \includegraphics[width=\linewidth]{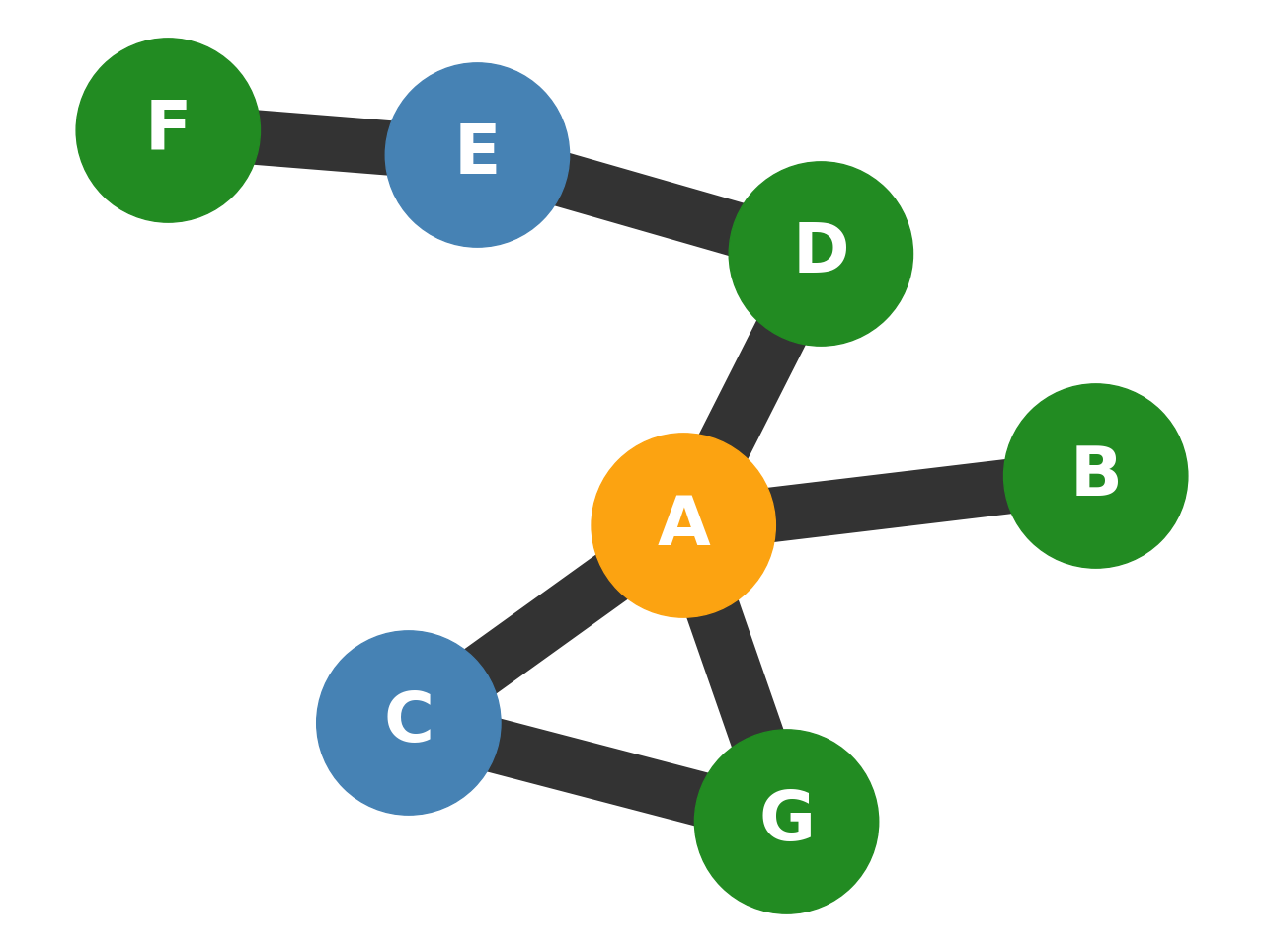}
        \caption{Step 1: Identify target node A}
        \label{fig:neighbor_a_opt}
    \end{subfigure}
    \hspace{2em} % Changed \hfill to \hspace{2em} for a smaller, fixed gap
    \begin{subfigure}{0.3\linewidth}
        \includegraphics[width=\linewidth]{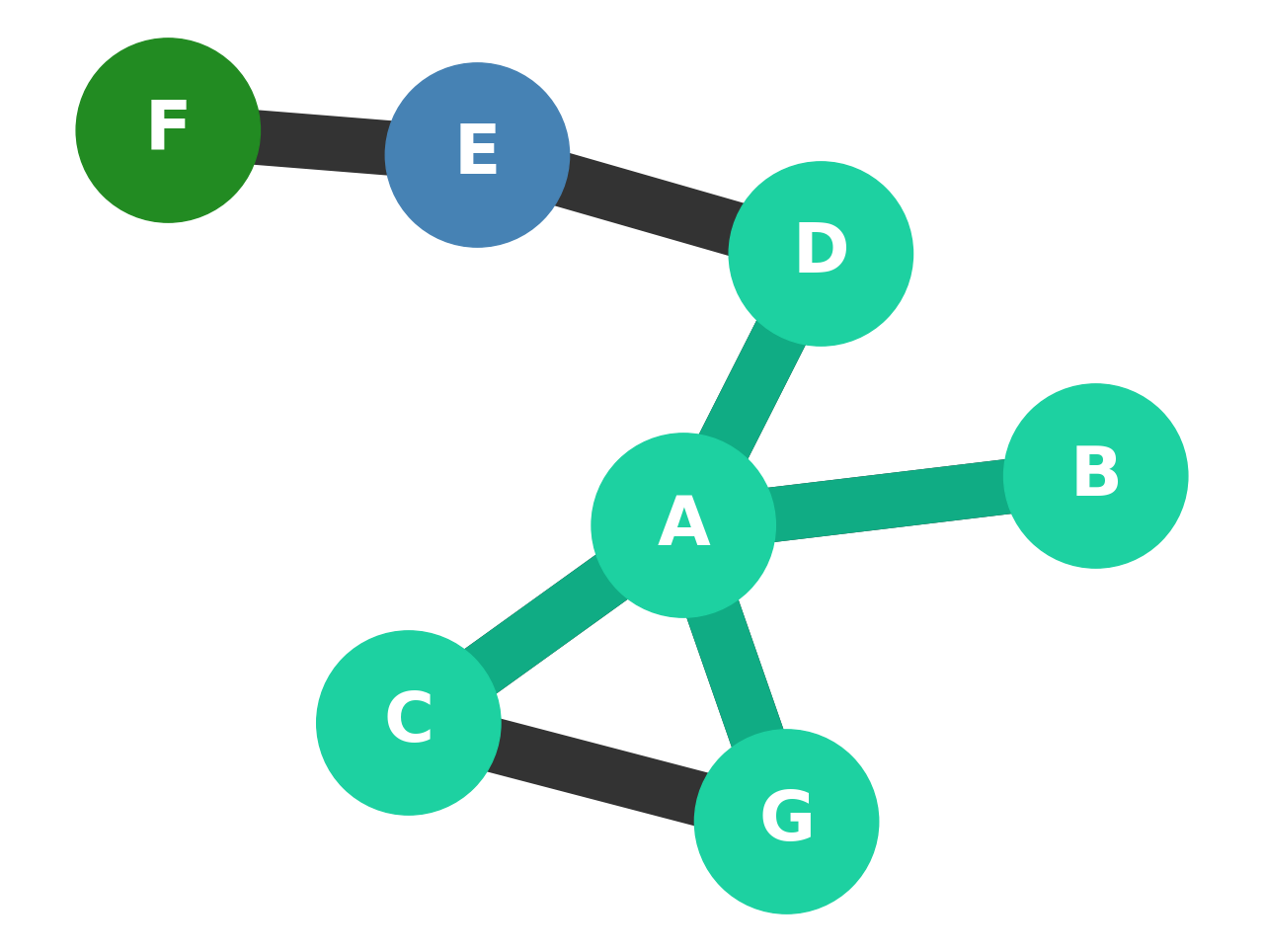}
        \caption{Step 2: Highlight direct neighbors}
        \label{fig:neighbor_b_opt}
    \end{subfigure}
    \caption{Visualizing \textbf{Neighbor Retrieval} for a target node A. The process identifies all nodes that are directly connected to node A.}
    \label{fig:neighbor_retrieval_optimized}
\end{figure*}

% \twocolumn % Switch back to two columns for the content
\section{Stability Analysis of DPO Training}\label{sec:stability_DPO}

To further verify the robustness of the Visual Reasoning Agent ($\mathcal{M}_{VRA}$) within GraphVista, we conducted an in-depth analysis of the training stability of Process-level Direct Preference Optimization (DPO). This section presents detailed training metrics for Qwen2.5-VL-7B fine-tuned using DPO.

\subsection{Training Loss and Reward Curves}

Figure \ref{fig:training_curves} illustrates the evolution of loss and implicit rewards over 7,500 training steps.

\begin{figure*}[t]
    \centering
    \begin{subfigure}[b]{0.48\textwidth}
        \centering
        \includegraphics[width=\linewidth]{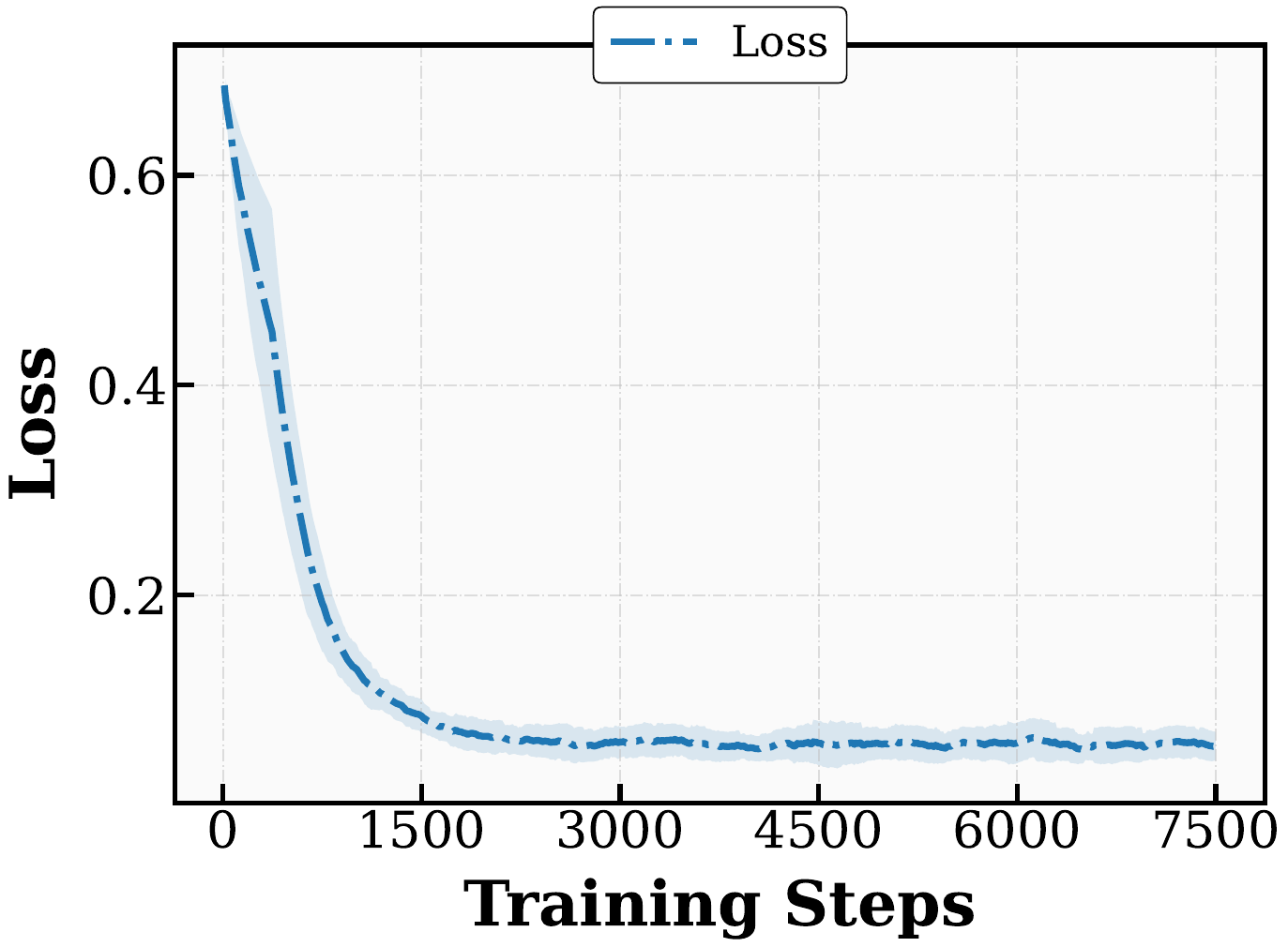}
        \caption{Training Loss}
        \label{fig:dpo_loss}
    \end{subfigure}
    \hfill
    \begin{subfigure}[b]{0.48\textwidth}
        \centering
        \includegraphics[width=\linewidth]{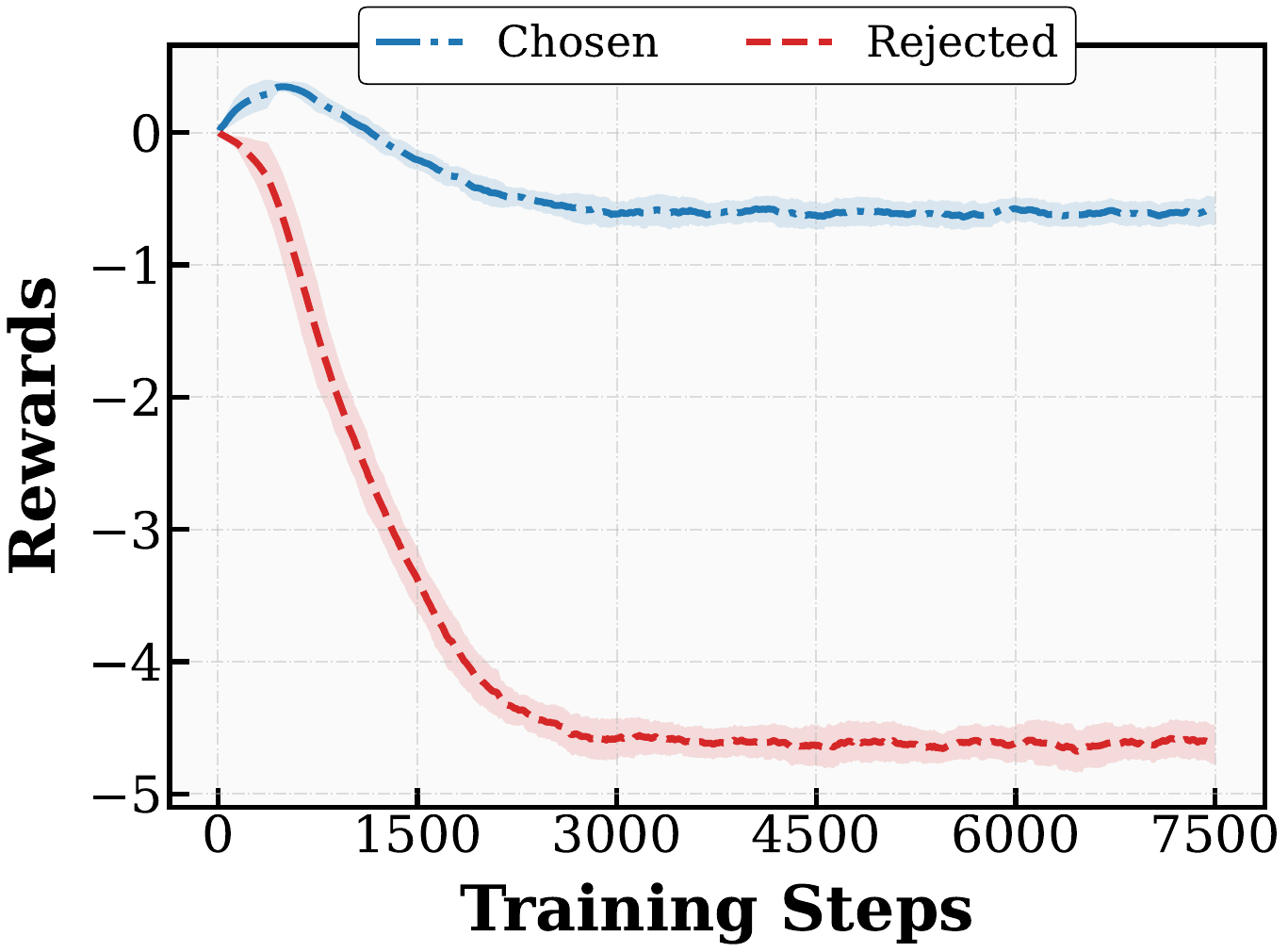}
        \caption{Implicit Rewards}
        \label{fig:dpo_rewards}
    \end{subfigure}
    \caption{DPO training dynamics of Qwen2.5-VL-7B on the Grena preference dataset. The left panel shows rapid convergence of training loss; the right panel demonstrates that the model successfully widened the reward margin between Chosen and Rejected paths.}
    \label{fig:training_curves}
\end{figure*}

\paragraph{Analysis of Training Dynamics}
The training process exhibited high stability without the divergence often observed in RL fine-tuning:

\begin{itemize}
    \item \textbf{Loss Convergence:} The training loss (Figure \ref{fig:dpo_loss}) displays a smooth downward trend, entering a stable convergence phase around 4,500 steps. This indicates that the model effectively learned the distribution of the preference data.
    
    \item \textbf{Reward Margin Separation:} As shown in Figure \ref{fig:dpo_rewards}, the reward values for \textit{Chosen} paths (indicated by the blue solid line) remain relatively stable, while the reward values for \textit{Rejected} paths (indicated by the orange dashed line) decrease significantly with training steps. This indicates that the optimizer maximizes the preference margin primarily by suppressing negative samples (i.e., reasoning chains containing visual hallucinations or logical errors), aligning with the theoretical expectations of DPO.
\end{itemize}

The observed stability is primarily attributed to our mild regularization strategy ($\beta=0.1$) and the high-quality process preference dataset derived from real topological structures.

\section{Example of Visual Graph Thought Chains}

This section provides concrete examples of the step-by-step "Visual Graph Thoughts" process, a core component of our framework for solving complex local reasoning tasks. Each figure-set deconstructs a graph-based problem into a sequence of visual states. This methodology allows the VLM to perform grounded reasoning by identifying and highlighting relevant nodes and edges at each step, culminating in a final solution.

Figure~\ref{fig:triangle_detection_optimized} illustrates the process for \textbf{Triangle Detection}. The chain begins by identifying the neighbors of a central node, then visually inspects for edges between those neighbors to confirm a 3-cycle. For \textbf{Shortest Path} finding, shown in Figure~\ref{fig:shortest_path_optimized}, the model emulates a breadth-first search, extending the path layer-by-layer from the source until the destination is reached. Finally, Figure~\ref{fig:neighbor_retrieval_optimized} depicts the fundamental operation of \textbf{Neighbor Retrieval}, where the model identifies and highlights all nodes directly adjacent to a given target.

%%%%%%%%%%%%%%

% --- Custom Colors ---
\definecolor{mainblue}{RGB}{44, 62, 80}
\definecolor{accentcyan}{RGB}{22, 160, 133}
\definecolor{lightgraybg}{RGB}{245, 247, 250}

% --- Custom Box Definitions ---

\newtcolorbox{categorybox}[1][]{
  enhanced,
  colback=white,
  colframe=mainblue,
  coltitle=white,
  fonttitle=\bfseries\sffamily\large,
  title={Task Categorization Examples},
  arc=3mm,
  boxrule=0.5pt,
  left=2mm, right=2mm, top=2mm, bottom=2mm, 
  drop shadow, 
  #1
}

\newtcolorbox{systempromptbox}[1][]{
  enhanced,
  colback=lightgraybg,
  colframe=black!75,
  colbacktitle=black!75,
  coltitle=white,
  fonttitle=\bfseries\sffamily,
  title={#1},
  arc=2mm,
  boxrule=0.8pt,
  left=4mm, right=4mm, top=4mm, bottom=4mm,
  drop shadow,
  attach boxed title to top left={xshift=5mm, yshift*=-\tcboxedtitleheight/2},
  boxed title style={
    colback=black!75,
    arc=1mm,
    boxrule=0pt,
    size=small
  },
  top=5mm 
}

\newtcolorbox{behaviorbox}[1]{
  enhanced,
  colback=white,
  colframe=accentcyan!80!black,
  coltitle=white,
  fonttitle=\bfseries\sffamily,
  title={#1},
  arc=3mm,
  boxrule=0.5pt,
  left=5mm, right=5mm, top=6mm, bottom=5mm, 
  drop shadow, 
  attach boxed title to top left={xshift=5mm, yshift*=-\tcboxedtitleheight/2},
  boxed title style={
    colback=accentcyan,
    colframe=accentcyan!80!black,
    arc=2mm,
    boxrule=0.5pt,
    size=small,
    shadow={2mm}{-2mm}{0mm}{black!20} 
  }
}
\begin{figure*}[ht]
\centering
\begin{categorybox}
    \setlength{\tabcolsep}{6pt}
    
    % --- Column 1: Text-Modality ---
    \begin{minipage}[t]{0.31\textwidth}
        \raggedright
        \centering
        \textbf{\textcolor{mainblue}{1. Text-Modality Tasks}} \\
        \small
        \textit{(Attribute, Statistic, Retrieval)}
        
        \vspace{1.5em}
        \raggedright
        \textbf{Query:} \\
        ``What is the weight of the edge between Node 5 and 9?''
        
        \vspace{0.8em}
        \textbf{Logic:} \\
        Targets discrete attributes. Visual parsing is redundant; direct GraphRAG retrieval is optimal.
    \end{minipage}%
    \hfill
    {\color{gray!50}\vrule width 0.8pt}
    \hfill
    % --- Column 2: Visual-Modality ---
    \begin{minipage}[t]{0.31\textwidth}
        \centering
        \textbf{\textcolor{mainblue}{2. Visual-Modality Tasks}} \\
        \small
        \textit{(Topological Reasoning)}
        
        \vspace{1.5em}
        \raggedright
        \textbf{Query:} \\
        ``Identify a cycle of length 3 involving Node A.''
        
        \vspace{0.8em}
        \textbf{Logic:} \\
        Requires tracing paths. Visual inspection exploits spatial recognition, superior to textual adjacency processing.
    \end{minipage}%
    \hfill
    % --- Divider ---
    {\color{gray!50}\vrule width 0.8pt}
    \hfill
    % --- Column 3: Collaborative ---
    \begin{minipage}[t]{0.31\textwidth}
        \centering
        \textbf{\textcolor{mainblue}{3. Collaborative Tasks}} \\
        \small
        \textit{(Global-Local Composition)}
        
        \vspace{1.5em}
        \raggedright
        \textbf{Query:} \\
        ``Find the highest degree node and list its 2-hop neighbors.''
        
        \vspace{0.8em}
        \textbf{Logic:} \\
        \textbf{Step 1 (Text):} Global search for target node.\\
        \textbf{Step 2 (Visual):} Local subgraph reasoning.
    \end{minipage}
\end{categorybox}
\vspace{-5pt}
\caption{Categorization guidelines used by the Planning Agent to route tasks based on structural dependency profiles.}
\label{fig:task_categorization}
\end{figure*}

\section{Task Parsing and Routing Mechanism}
\label{sec:task_parsing_routing_details}

This section delineates the decision-making logic of the Planning Agent, which serves as the semantic router within the GraphVista framework. Building upon the architectural overview, we detail the criteria and operational rationale for allocating tasks across Text, Visual, and Collaborative modalities.

To balance computational efficiency with semantic flexibility, the Planning Agent employs a hierarchical \textbf{two-stage routing protocol}:

\begin{enumerate}
    \item \textbf{Deterministic Template Routing (Primary Strategy):} The agent initially queries a library of predefined task templates (e.g., \textit{``What is the degree of node $<$ID$>$?''}). Upon identifying an exact match, the task is deterministically typed and immediately routed to the designated modality, bypassing complex inference to minimize latency for standardized queries.
    \item \textbf{Semantic Inference Fallback (Secondary Strategy):} In the absence of a template match, the agent resorts to semantic parsing. This mechanism analyzes the \textit{structural dependency} of the query to infer user intent, specifically determining whether the solution necessitates topological traversal (Visual), attribute retrieval (Text), or a synergistic combination of both (Collaborative).
\end{enumerate}

\paragraph{Rationale for Text-Modality Assignment.}
The Text Modality is explicitly designated for tasks akin to database retrieval. This category encompasses: (1) \textit{Existence Verification}, such as confirming the presence of nodes or edges; (2) \textit{Attribute Retrieval}, which involves extracting specific properties (e.g., weights, labels) from the knowledge base $\mathcal{K}$; and (3) \textit{Global Aggregation}, for operations like degree counting. For these tasks, direct retrieval offers computational superiority over visual processing, effectively mitigating latency and eliminating the noise associated with rendering dense graphs for simple factual extraction.

\paragraph{Rationale for Visual-Modality Assignment.}
The Visual Modality is reserved for tasks necessitating "topological perception." Large Language Models (LLMs) processing linearized graph descriptions (e.g., adjacency lists) are prone to contextual attenuation ("lost-in-the-middle" phenomena) when tracing extended paths. Conversely, the visual modality projects the graph into a 2D layout where topological features—such as clusters, bridges, and cycles—become explicitly salient. This is critical for tasks like \textit{Shortest Path} or \textit{Motif Detection}, where the VLM can perceive connectivity patterns more effectively than through iterative textual deduction.

\paragraph{Rationale for Collaborative-Modality Assignment.}
Complex graph reasoning often demands a hybrid "Search-then-Reason" paradigm. Collaborative tasks are characterized by a \textit{global search space} (optimally handled by GraphRAG indexing) coupled with \textit{local structural verification} (optimally handled by Visual Graph Thoughts). The Planning Agent decomposes such queries into sequential sub-goals. For instance, in diameter estimation, the Text branch first filters peripheral node candidates to reduce the search space, after which the Visual branch executes fine-grained pathfinding on the induced subgraph.

\subsection{Analysis of Task Classification Accuracy}
\label{subsec:task_cls_analysis}

To evaluate the reliability of the routing mechanism, we quantify the task classification error rates across different VLM backbones. Table~\ref{tab:task_cls_error_rate} presents the category misclassification rates for the Planning Agent.
\begin{table}[ht]
\centering
\resizebox{\linewidth}{!}{
    \begin{tabular}{l c}
    \toprule
    \textbf{VLM} & \textbf{Error Rate (\%)} \\
    \midrule
    GLM-4.1V-9B         & 0.1650 \\
    InternVL3-9B        & 0.1593 
\\
    Qwen2.5-VL-7B       & 0.2482 \\
    Gemma-3-12B         & 0.1697 
    \\
    Qwen2.5-VL-7B (DPO) & 0.2495 \\
    Gemma-3-12B (DPO)   & 0.1713 \\
    \bottomrule
    \end{tabular}
}
\caption{Task Classification Error Rates across different VLM backbones. Models fine-tuned with DPO are denoted with (DPO).}
\label{tab:task_cls_error_rate}
\end{table}
The results highlight two key observations regarding the routing stability:
\begin{itemize}
    \item \textbf{Impact of Model Scale:} There is a clear correlation between model capacity and routing accuracy. Larger models, such as InternVL3-9B and GLM-4.1V-9B, consistently achieve lower error rates (approx. 0.16\%) compared to the 7B-parameter baselines (approx. 0.25\%). This suggests that the semantic nuance required for accurate intent classification benefits significantly from stronger foundational language understanding.
    \item \textbf{Robustness to DPO Fine-tuning:} The application of Direct Preference Optimization (DPO)—while crucial for enhancing the step-by-step visual reasoning detailed in the main methodology—does not degrade the semantic routing capability. The error rates for DPO-tuned models (e.g., Qwen2.5-VL-7B (DPO) at 0.2495) remain comparable to their base counterparts (0.2482). This indicates that our alignment strategy effectively isolates visual reasoning improvements without incurring an ``alignment tax'' on the model's general instruction-following and planning abilities.
\end{itemize}

\begin{figure*}[ht]
    \centering
    \begin{systempromptbox}[System Prompt for Planning Agent Task Parsing and Routing]
    \footnotesize 
    \textbf{Role Definition:}
    You are an expert Graph Analysis Planning Agent. Your objective is to parse user queries regarding graph structures, extract key entities, classify the specific task type, and route the task to the optimal processing modality.
      
    \vspace{0.2cm}
    \textbf{Parsing Logic:}
    \begin{enumerate}
        \item \textbf{Template Verification:} Initially, verify if the user input aligns with any predefined templates (e.g., Degree, Weight, Shortest Path). If a match is detected, output the standard JSON immediately.
        \item \textbf{Semantic Fallback:} Upon template mismatch, analyze the "structural dependency" of the request to determine the required modality (Text vs. Visual vs. Collaborative).
    \end{enumerate}

    \vspace{0.2cm}
    \textbf{Task Categorization Guidelines (for Fallback):}
    \begin{itemize}
        \setlength{\itemsep}{1pt}
        \item \textbf{Text-Modality Tasks:} Tasks requiring direct retrieval of statistical data, node attributes, or global counts devoid of complex structural reasoning.
        \textit{Examples: Node Count, Node Degree, Edge Existence, Edge Weight.}
        \item \textbf{Visual-Modality Tasks:} Tasks necessitating topological understanding, pathfinding, or pattern recognition within a local subgraph.
        \textit{Examples: Shortest Path, Cycle/Triangle Detection, Common Neighbors, Clique Detection.}
        \item \textbf{Modality-Collaborative Tasks:} Tasks requiring global structural understanding necessitating decomposition into sequential text and visual sub-tasks.
        \textit{Examples: Graph Diameter, Critical Node Detection, Connectivity Analysis.}
    \end{itemize}
      
    \vspace{0.2cm}
    \textbf{Output Schema (JSON):}
    \vspace{-0.2cm}
    \begin{verbatim}
{
  "task_type": "String (e.g., shortest_path, node_degree)",
  "entities": ["List of key node IDs or null"],
  "modality": "One of [Text, Visual, Collaborative]",
  "reasoning": "Brief explanation of the modality choice",
  "decomposition": ["List of sub-steps if modality is Collaborative, else null"]
}
    \end{verbatim}
    \vspace{-0.2cm}
      
    \textbf{Few-Shot Demonstrations:}
    \vspace{-0.1cm}
      
    \textit{User Input:} "What is the degree of node 15?"
    \textit{Agent Output:}
    \vspace{-0.2cm}
    \begin{verbatim}
{
  "task_type": "node_degree", "entities": ["15"], "modality": "Text",
  "reasoning": "Template Match: 'node_degree'. Retrieves attribute via GraphRAG.",
  "decomposition": null
}
    \end{verbatim}
    \vspace{-0.2cm}
      
    \textit{User Input:} "Find the shortest path between node A and node B."
    \textit{Agent Output:}
    \vspace{-0.2cm}
    \begin{verbatim}
{
  "task_type": "shortest_path", "entities": ["A", "B"], "modality": "Visual",
  "reasoning": "Template Match: 'shortest_path'. Topology required.",
  "decomposition": null
}
    \end{verbatim}
    \vspace{-0.2cm}
      
    \textit{User Input:} "Analyze the network to find bottlenecks impacting flow."
    \textit{Agent Output:}
    \vspace{-0.2cm}
    \begin{verbatim}
{
  "task_type": "bottleneck_detection", "entities": [], "modality": "Collaborative",
  "reasoning": "Fallback: Unmatched template. Semantic parsing indicates global 
  search followed by local verification.",
  "decomposition": [
      "Identify high-betweenness candidates (Text)",
      "Analyze local connectivity for flow restriction (Visual)"
  ]
}
    \end{verbatim}
      
    \textbf{Current Request:} \texttt{<USER\_QUERY>}
    \end{systempromptbox}
    \caption{The system prompt utilized by the Planning Agent. The mechanism prioritizes template matching for efficiency, falling back to semantic parsing for complex or unstructured queries.}
    \label{fig:planning_agent_prompt}
\end{figure*}

\section{Prompt and QA Templates}
\label{sec:prompt_templates}

This section delineates the standardized prompt templates and Question-Answering (QA) protocols utilized to rigorously evaluate the graph structure understanding capabilities of Vision-Language Models (VLMs) within the proposed \textbf{GraphVista} framework. To ensure a comprehensive assessment of both retrieval accuracy and reasoning depth, we categorize these templates into \textit{Text-Modality Tasks} and \textit{Visual-Modality Tasks}. This categorization strictly adheres to the task routing logic executed by the Planning Agent, as detailed in Section~3.3 of the main paper.

\subsection{Text-Modality Tasks: Retrieval via GraphRAG}
\label{subsec:text_modality_prompts}

The Text-Modality category encompasses tasks necessitating precise information retrieval and statistical aggregation rather than abstract topological reasoning. Upon classifying a query as text-dependent, the Planning Agent routes the task to the text-modality branch. This branch leverages the \textbf{Hierarchical GraphRAG Base ($\mathcal{K}$)} to execute targeted retrieval operations. By accessing the structured textual descriptions stored within $\mathcal{K}$ (e.g., node attributes, edge weights, and local neighborhoods), the model extracts factual information directly, thereby bypassing the potential ambiguity associated with visual inference for non-spatial queries.

\subsection{Visual-Modality Tasks: Reasoning with Visual Graph Thoughts}
\label{subsec:visual_modality_prompts}

This category targets tasks demanding intricate topological perception and multi-step logical inference, such as pathfinding or cycle detection. For such queries, the Planning Agent delegates execution to the visual-modality branch. This module orchestrates a pipeline consisting of: (1) \textit{dynamic subgraph extraction} from $\mathcal{K}$ to isolate the region of interest; (2) \textit{task-driven visualization} to render the subgraph into a high-resolution image; and (3) the deployment of the \textbf{Visual Graph Thoughts Agent}. This agent employs a chain-of-thought mechanism grounded in the generated visual evidence to perform iterative, step-by-step reasoning, ensuring that the final answer is topologically consistent.

\begin{figure*}[ht]
\centering
\begin{behaviorbox}{Task 1: Node Existence and Properties}
\textbf{Prompt Template:}  
\textit{"Determine if node \texttt{<node\_id>} exists in the graph. If so, what is its degree?"}

\vspace{0.8em}
\textbf{GraphVista Behavior:}  
The planning agent categorizes this as a text-modality task. The model performs a targeted retrieval query against the GraphRAG Base $\mathcal{K}$ to find the entry for \texttt{<node\_id>}. The retrieved context, containing the node's 1-hop neighborhood information, is then used to generate a direct answer.

\vspace{0.8em}
\textbf{Rationale:}  
This task involves retrieving explicitly stored attributes. The textual representation in the GraphRAG base is optimized for such direct lookups, making it efficient and reliable.
\end{behaviorbox}
\end{figure*}

\begin{figure*}[ht]
\centering
\begin{behaviorbox}{Task 2: Edge Existence and Weight}
\textbf{Prompt Template:}  
\textit{"Is there a direct edge between node \texttt{<node\_id\_1>} and node \texttt{<node\_id\_2>}? If yes, provide its weight."}

\vspace{0.8em}
\textbf{GraphVista Behavior:}  
The query is identified as a text-modality task. The system retrieves the adjacency information for \texttt{<node\_id\_1>} from $\mathcal{K}$. The VLM then processes this structured text to check for a connection to \texttt{<node\_id\_2>} and extracts the corresponding weight attribute.

\vspace{0.8em}
\textbf{Rationale:}  
Like node property lookups, edge verification relies on retrieving specific, pre-processed information. The textual modality avoids the potential for visual ambiguity (e.g., overlapping edges or nodes) that could occur in a complex graph visualization.
\end{behaviorbox}
\end{figure*}

\begin{figure*}[ht]
\centering
\begin{behaviorbox}{Task 3: Shortest Path Identification}
\textbf{Prompt Template:}  
\textit{"Find and list the sequence of nodes that form the shortest path between node \texttt{<node\_id\_1>} and node \texttt{<node\_id\_2>}."}

\vspace{0.8em}
\textbf{GraphVista Behavior:}  
The planning agent categorizes this as a visual-modality task. The Visual Graph Thoughts Agent extracts a multi-centric subgraph centered around the shortest path candidates between \texttt{<node\_id\_1>} and \texttt{<node\_id\_2>}. This high-resolution subgraph is visualized. The VLM then initiates a Visual Graph Thought process, highlighting nodes and edges step-by-step in the image to trace the optimal path, verbalizing its reasoning at each step before presenting the final sequence.

\vspace{0.8em}
\textbf{Rationale:}  
Textual descriptions of paths can be convoluted and difficult for VLMs to reason over effectively. A visual representation provides a “what-you-see-is-what-you-get” view of the topology, which is more intuitive for complex structural reasoning and naturally supports stepwise inference grounded in visual evidence.
\end{behaviorbox}
\end{figure*}

\begin{figure*}[ht]
\centering
\begin{behaviorbox}{Task 4: Triangle Detection}
\textbf{Prompt Template:}  
\textit{"Does a 3-cycle (triangle) involving node \texttt{<node\_id>} exist? If so, list the nodes of one such triangle."}

\vspace{0.8em}
\textbf{GraphVista Behavior:}  
This is routed to the visual modality. An ego-centric subgraph around \texttt{<node\_id>} is extracted from $\mathcal{K}$ and visualized. The Visual Graph Thoughts agent first identifies the neighbors of \texttt{<node\_id>} in the image. It then visually inspects for edges connecting any two of these neighbors. If such an edge is found, the VLM confirms the existence of a triangle and outputs the three nodes forming the cycle.

\vspace{0.8em}
\textbf{Rationale:}  
Detecting local structural patterns like cycles is inherently a topological problem. Visual inspection allows the VLM to directly perceive the connectivity pattern, whereas a text-based approach would require inefficient and error-prone traversal of adjacency lists to check for path closures.
\end{behaviorbox}
\end{figure*}

\begin{figure*}[ht]
\centering
\begin{behaviorbox}{Task 5: Graph Diameter Calculation (Modality-Collaborative)}
\textbf{Prompt Template:}  
\textit{"Calculate the diameter of the graph. First identify potential peripheral nodes, and then determine the maximum shortest path distance between them."}

\vspace{0.8em}
\textbf{GraphVista Behavior:}  
The Planning Agent identifies this as a collaborative task and decomposes it into two sequential sub-tasks. 
\begin{itemize}
    \item \textbf{Phase 1 (Text Modality):} The agent queries the GraphRAG Base $\mathcal{K}$ to identify a set of "peripheral nodes" (typically defined as nodes with low centrality scores in Tier 3). This filters the search space to the most likely candidates for the diameter endpoints.
    \item \textbf{Phase 2 (Visual Modality):} The system extracts multi-centric subgraphs connecting these candidate pairs. The Visual Graph Thoughts agent then visually traces the shortest paths between them to calculate their distances. 
\end{itemize}
Finally, the system aggregates these results and returns the maximum distance found as the graph diameter.

\vspace{0.8em}
\textbf{Rationale:}  
Calculating diameter requires both global search and local pathfinding. The text modality is efficient for globally filtering candidate nodes based on statistical properties (centrality), avoiding the need to process the entire graph visually. Conversely, the visual modality excels at the subsequent topological reasoning required to accurately count path lengths between specific pairs, which is cumbersome for text-based reasoning.
\end{behaviorbox}
\end{figure*}

\end{document}